\documentclass[sigconf]{acmart}
\AtBeginDocument{%
  }

\setcopyright{acmlicensed}
\copyrightyear{2018}
\acmYear{2018}
\acmDOI{XXXXXXX.XXXXXXX}
\acmConference[Conference acronym 'XX]{Make sure to enter the correct
  conference title from your rights confirmation email}{June 03--05,
  2018}{Woodstock, NY}
\acmISBN{978-1-4503-XXXX-X/2018/06}

\PassOptionsToPackage{table,dvipsnames}{xcolor}  
\usepackage{xcolor}  
\usepackage{colortbl}  

\usepackage{enumitem}
\usepackage{amsthm}
\usepackage{hyperref} 
\usepackage{multirow}
\usepackage{multicol}
\usepackage{tcolorbox}
\usepackage{graphicx}
\usepackage{caption}
\usepackage{subcaption}

\newcounter{ObservationCounter}
\newcommand{\Obs}[1]{
  \refstepcounter{ObservationCounter}
  \noindent\textbf{\underline{Observation \theObservationCounter:}} #1
}

\newcommand{\cred}[1]{\color{red}{\textbf{#1}}}
\newcommand{\cblue}[1]{\color{blue}{\textbf{#1}}}

\begin{document}

\title{Exploring the Potential of Large Language Models as Predictors in Dynamic Text-Attributed Graphs}

\author{Runlin Lei}
\affiliation{%
  \institution{Renmin University of China}
  \city{Beijing}
  \country{China}}
\affiliation{%
  \institution{Ant Group}
  \city{Beijing}
  \country{China}}
\email{runlin_lei@ruc.edu.cn}

\author{Jiarui Ji}
\affiliation{%
  \institution{Renmin University of China}
  \city{Beijing}
  \country{China}}
\email{2023100839@ruc.edu.cn}

\author{Haipeng Ding}
\affiliation{%
  \institution{Renmin University of China}
  \city{Beijing}
  \country{China}}
\email{dinghaipeng@ruc.edu.cn}

\author{Lu Yi}
\affiliation{%
  \institution{Renmin University of China}
  \city{Beijing}
  \country{China}}
\email{yilu@ruc.edu.cn}

\author{Zhewei Wei}
\affiliation{%
  \institution{Renmin University of China}
  \city{Beijing}
  \country{China}}
  \authornote{Zhewei Wei and Yongchao Liu are the corresponding authors. }
\email{zhewei@ruc.edu.cn}

\author{Yongchao Liu}
\affiliation{%
  \institution{Ant Group}
  \city{Hangzhou}
  \country{China}}
  \authornotemark[1]
\email{yongchao.ly@antgroup.com}

\author{Chuntao Hong}
\affiliation{%
  \institution{Ant Group}
  \city{Beijing}
  \country{China}}
\email{chuntao.hct@antgroup.com}

\begin{abstract}
With the rise of large language models (LLMs), there has been growing interest in Graph Foundation Models (GFMs) for graph-based tasks.
By leveraging LLMs as predictors, GFMs have demonstrated impressive generalizability across various tasks and datasets.
However, existing research on LLMs as predictors has predominantly focused on static graphs, leaving their potential in dynamic graph prediction unexplored.
In this work, we pioneer using LLMs for predictive tasks on dynamic graphs. 
We identify two key challenges: the constraints imposed by context length when processing large-scale historical data and the significant variability in domain characteristics, both of which complicate the development of a unified predictor.
To address these challenges, we propose the GraphAgent-Dynamic (GAD) Framework, a multi-agent system that leverages collaborative LLMs.  
In contrast to using a single LLM as the predictor, GAD incorporates global and local summary agents to generate domain-specific knowledge, enhancing its transferability across domains. 
Additionally, knowledge reflection agents enable adaptive updates to GAD's knowledge, maintaining a unified and self-consistent architecture.
In experiments, GAD demonstrates performance comparable to or even exceeds that of full-supervised graph neural networks without dataset-specific training.
Finally, to enhance the task-specific performance of LLM-based predictors, we discuss potential improvements, such as dataset-specific fine-tuning to LLMs.  
By developing tailored strategies for different tasks, we provide new insights for the future design of LLM-based predictors.
\end{abstract}


\received{20 February 2007}
\received[revised]{12 March 2009}
\received[accepted]{5 June 2009}

\maketitle

\section{Introduction}
Graphs are ubiquitous in real-world applications, serving as a fundamental data structure for modeling interactions between entities~\cite{Application_GNN, recommandation_survey, DGNN_survey}.
In practice, a significant portion of graphs are Dynamic Text-Attributed Graphs (DyTAGs), where nodes and edges are enriched with textual attributes and evolve over time.  
A typical example is social networks, where nodes (representing users) contain textual profiles or descriptions, and interactions between users are primarily text-based. 
Over time, users form new interactions, and new users are introduced to the network.

To effectively capture both textual and structural information, integrating large language models (LLMs) into the pipeline for TAGs has become a prominent research focus, particularly with LLMs-as-predictors~\cite{InstructGLM_23, GraphGPT_24, UniGraph_24, OneForAll_24, GOFA_24}. 
These methods contribute to the unification of various downstream tasks by leveraging textual information in graphs.
Compared to using Graph Neural Networks (GNNs) for prediction, employing LLMs as predictors offers the following advantages:  
(1) LLMs demonstrate strong transferability across tasks and datasets, enabling adaptation to new scenarios without retraining. In contrast, GNNs require task- and dataset-specific training, limiting their flexibility in dynamic environments.  
(2) LLMs produce more comprehensible predictions by providing reasoning insights, whereas GNNs often function as black-box models.  
(3) LLMs are inherently suited for generative tasks, whereas GNNs face fundamental challenges in this area.  

Despite the above advantages, existing research on LLM-based predictors has primarily focused on static graphs, where nodes and edges remain unchanged over time.  
Incorporating the time dimension, dynamic graphs introduce greater challenges in designing LLM-based predictors due to their increased complexity.  
Existing studies about applying LLMs to dynamic graphs mainly focus on small-scale temporal graph reasoning tasks~\cite{LLM4DyG_24, Test_of_Time_24}.
These investigations are limited to non-textual-attributed graphs, focus solely on temporal reasoning, and rely on datasets that are too small to reflect real-world scenarios.  
A recent study, the Dynamic Text-Attributed Graph Benchmark (DTGB)~\cite{DTGB_2024}, introduces LLMs to address the future text generation task in DyTAGs. 
However, the authors still rely on GNNs as predictors for predictive tasks, inheriting their limitations compared to LLM-based predictors.  
Based on these observations, the following question arises: \textit{Can LLMs serve as strong predictors in DyTAGs as well?}

\begin{figure*}[t]
  \centering
  \includegraphics[width=\textwidth]{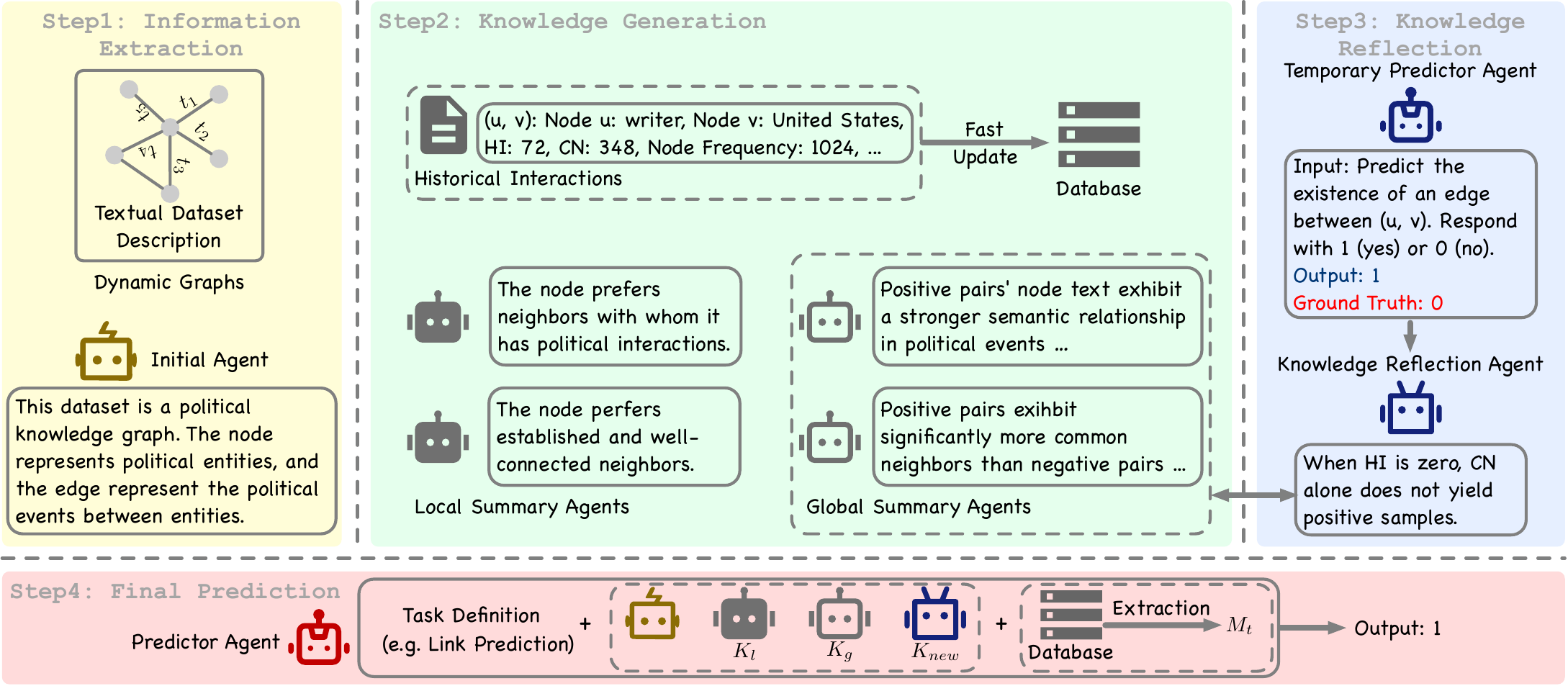}
  \caption{Overview of the GAD pipeline. The Predictor Agent utilizes generated knowledge and relevant metrics to make predictions. The database is updated over time, while the knowledge is reused unless updated through reflection.}
  \label{fig:framework}
  \vspace{-1.5em}
\end{figure*}

In this paper, we pioneer the exploration of LLMs as predictors for predictive tasks in dynamic graphs. 
For the three predictive tasks in DTGB, we first explore single \texttt{text-only} and \texttt{structure-aware} LLM-based predictors, mirroring the attempts made in static graphs~\cite{LLM_explor_Zhikai_23}. 
We identify new challenges for LLM-based predictors arising from the unique characteristics of DyTAGs.  
Firstly, the substantial volume of historical interactions imposes significant constraints on the context window, often exceeding length limits and hindering the effectiveness of LLM-based predictors.  
Moreover, dataset variability makes it challenging for a universal prompt to generalize effectively.  
To address these challenges, we design a multi-LLM-based agent collaboration framework, GraphAgent-Dynamic (GAD).
The workflow of GAD is displayed in Figure~\ref{fig:framework}.
Specifically, we employ an Initial Agent to extract dataset and task description, Global Summary Agents to summarize task- and dataset-specific knowledge, Local Summary Agents to capture fine-grained local preferences, and Knowledge Reflection Agents to provide knowledge supplementation based on prediction trajectory. 
Finally, we use a Prediction Agent to generate the final prediction.   
GAD only requires a predefined template and human-written dataset description as input and can perform various downstream tasks on different datasets through a unified framework. 
Experimental results demonstrate that GAD can perform comparable or even superior to GNNs without training.
Finally, to enhance LLM-based predictors in DyTAGs further, we explore task-specific and dataset-specific strategies for improvement.
Our contributions are as follows:
\begin{itemize}[leftmargin=*]
    \item We pioneer the exploration of LLMs as predictors for DyTAGs, considering both \texttt{text-only} and \texttt{structure-aware} variants.
    Our results demonstrate that a single LLM as the predictor can achieve performance comparable to GNNs in specific tasks.  
    \item We identify the key challenge for single-LLM-based predictors in DyTAGs: the significant variability across domains where unified knowledge fails to guide the LLM effectively. 
    To address this, we propose GAD, a multi-agent framework based on collaborative LLMs for DyTAGs.
    Experimental results demonstrate its superior generalizability over single-LLM-based predictors.
    \item We explore targeted improvement strategies for LLM-based predictors, including domain-specific fine-tuning of LLMs and the design of domain-specific recallers. 
    In particular, we identify the inherent bottleneck in the future edge classification task.
    These insights motivate future designs of LLM-based predictors. 
\end{itemize}

\section{Related Work}
\textbf{LLMs for Temporal Tasks}
The application of LLMs to temporal data has garnered significant research attention in recent years. 
\citet{TEST_24} and \citet{LLM4TS_23} explore the use of LLMs in time series analysis, demonstrating their capability to model complex temporal patterns.
In the domain of dynamic graphs, ~\citet{Test_of_Time_24} investigate the ability of LLMs to perform temporal reasoning tasks over graph structures, highlighting their potential to capture temporal dependencies. 
Similarly, ~\citet{LLM4DyG_24} study the effectiveness of LLMs in understanding spatial-temporal relationships within graphs. 
However, these studies focus primarily on temporal reasoning tasks in small-scale settings rather than real-world predictive tasks involving large-scale graphs. \\
\textbf{LLMs for DyTAGs}
Recently, there have been initial efforts to investigate the use of LLMs for DyTAGs. 
\citet{Temporal_GFM} study the neural scaling laws of temporal graph foundation models through pretraining on a collection of temporal graphs. 
However, their work is constrained by the limited variety of dataset domains and focuses exclusively on a single graph property prediction task.
\citet{DTGB_2024} propose the first DyTAG benchmark, incorporating LLMs for the future edge text generation task. 
Nonetheless, their approach to prediction tasks still relies on shallow text embeddings combined with GNNs instead of LLMs as predictors.
Overall, the potential of LLMs as predictors in DyTAG tasks remains largely unexplored compared to those in static TAGs.

More related works about LLM-empowered Agents and LLMs for static TAGs are provided in Appendix~\ref{app:related works}.
\section{A Single LLM as the Predictor}
In static TAGs, LLM-based predictors have demonstrated superior performance and generalizability across various tasks and datasets, leveraging a unified model or framework~\cite{LLM_explor_Zhikai_23,InstructGLM_23, LLaGa_24, UniGraph_24, GOFA_24}. 
Building on these studies, we start by employing a single vanilla LLM as the predictor to perform prediction tasks in DyTAGs. 

\subsection{Preliminaries}
\textbf{DyTAGs.} In this paper, we focus on Continuous-Time Dynamic Graphs (CTDGs), following studies~\cite{DTGB_2024, DygFormer_23, TGBSeq}. 
Formally, we denote a CTDG as $\mathcal{G}=(\mathcal{V},\mathcal{E})$, where the edge set $\mathcal{E}$ is denoted as a stream of timestamped edges, i.e., 
$\mathcal{E} = \{(u_0, v_0, t_0), \dots, (u_{T}, v_{T}, t_{T})\}$ with $t_0\le t_1\le\ldots\le t_{T}$. An edge $(u_i, v_i, t_i)$ denotes that the source node $u_i\in\mathcal{V}$ and the destination node $v_i\in \mathcal{V}$ interact at time $t_i$. 
DyTAGs are dynamic graphs that include textual descriptions of nodes, interactions, and edge labels, which can be commonly seen in real-world applications. 
Following DTGB~\cite{DTGB_2024}, we define the textual description of a node \( v \) as \( d_v \), the edge description between source node \( u \) and destination node \( v \) at timestamp \( t \) as \( r_{u,v,t} \), and the corresponding textual edge category as \( c_{u,v,t} \).
At time \( t \), the historical neighbor set of node \( u \), denoted as \( \mathcal{N}_t(u) \), is defined as:
\[
\mathcal{N}_t(u) = \{v \mid (u, v, t_p) \in \mathcal{E}, t_p < t\}
\]
which includes all nodes \( v \) that have interacted with \( u \) at any prior timestamp \( t_p \).
Additionally, we define \( p_t(u,v) \) as the distribution of edge labels for the interactions between \( u \) and \( v \) before timestamp \( t \), which is defined as:
\[
p_t(u,v) = \{c_1: n_1, c_2: n_2, \dots, c_k: n_k\}
\]
where \( c_i \) represents the edge label of the \( i \)-th class and \( n_i \) is the count of occurrences of label \( l_i \) prior to \( t \).
We define $p_t(u)$ similarly, but to describe the distribution of edge labels for the interactions $u$ historically involved in. \\
\textbf{LLM-based Predictors.}  
Denote the LLM-based predictor as \( f \), the prediction process at timestamp $t$ is described as:
\begin{equation}
M_t = g(\mathcal{G}), \quad P = f(M_t, K), \label{eq:1}
\end{equation}
where \( g(\cdot) \) is an extraction function that retrieves the relevant information \( M_t \) from the graph \( \mathcal{G} \), and \( K \) represents the knowledge that guides the LLM to correctly utilize \( M_t \).
A unified LLM-based predictor generates accurate predictions \( P \) \textbf{without additional training when transferred to new tasks or datasets}.

\subsection{General Experiment Setup}
We outline general settings for all the experiments here. 
Unless otherwise stated, the settings will be followed throughout the paper. \\
\textbf{Datasets.} We use five datasets Enron, GDELT, Googlemap\_CT, ICEWS1819, and Stack\_elec from DTGB~\cite{DTGB_2024}.
These datasets cover four different domains, with at least 6,786 nodes and 797,907 interactions. 
The statistics are provided in Appendix~\ref{app:data_stats}.\\
\textbf{Tasks.} We consider all three predictive tasks from DTGB~\cite{DTGB_2024}: future link prediction (LP), node retrieval (NR), and future edge classification (EC).
In future link prediction, we aim to predict whether a link will form between nodes \((u, v)\) at future timestamps based on historical data. 
In node retrieval, the goal is to rank candidate destination nodes, consisting of one positive sample and one hundred negative samples, by their likelihood of interacting with a source node \( u \) in the future. 
In future edge classification, we aim to predict the class of future edges between nodes \((u, v)\) using only historical data without edge attributes.
More details of task description are provided in Appendix~\ref{app:task_description}. \\
\textbf{Implementation Details.}
For GNN baselines, we select TCL~\cite{TCL_21}, GraphMixer~\cite{GraphMixer_23}, and DyGFormer~\cite{DygFormer_23} from DTGB.
These models are advanced and efficient, while other models either exceeded one day per run in training or encountered out-of-memory issues in our experiments. 
We choose the best parameters provided by DTGB and use Bert embeddings for initialization. 
We consider the transductive setting with random negative sampling since it is the most basic evaluation setting.
We include three LLM backbones: DeepSeek-V3 (referrd to as DeepSeek)~\cite{deepseekai2024deepseekv3technicalreport}, GPT4o-mini-0718 (referred to as GPT)~\cite{openai2024gpt4mini}, and Llama-3-8b (referred to as Llama)~\cite{llama3modelcard}. 
If the name of LLM is not mentioned in subsequent sections, we default to DeepSeek as the backbone. 
\\
\textbf{Evaluation Protocol.}
Following previous works~\cite{GraphMixer_23, DygFormer_23, DTGB_2024, TGBSeq}, we chronologically split each dataset into train/validation/test sets with a ratio of 70\%/15\%/15\%. 
Considering the high cost of LLM inference, we sample 10,240 samples that are chronologically closest to the validation set from the test set (40 batches with a batch size of 256), ensuring temporal continuity in the dataset.

For the evaluation metrics, we use accuracy for future link prediction, Hits@k for node retrieval, and weighted Precision/Recall/F1 score for future edge classification. 
These metrics are consistent with those used in DTGB, except that we do not use Average Precision and AUC-ROC for future link prediction. 
This is because the output of LLMs is in text form rather than logits, making these calculations infeasible.\\
\textbf{Dataset Description Extraction.}
We incorporate an Initial Agent to provide the dataset description. 
Using the human-written dataset description from DTGB \cite{DTGB_2024} as input, the Initial Agent extracts and summarizes the physical meanings of \textit{graph}, \textit{node}, and \textit{edge}, as well as the corresponding node and edge texts.
The extracted message is included in the prompt for all LLMs.~\footnote{The term 'single' in this section refers to the exclusive use of the Predictor Agent for downstream tasks alongside the Initial Agent.}

\subsection{Results of a Single LLM as the Predictor}
\label{Sec:given text}
In the static TAG node classification task, a single LLM as the predictor has been shown to achieve performance comparable to, or even surpass, specialized fully supervised GNNs \cite{LLM_explor_Zhikai_23}. 
In the PubMed dataset, the zero-shot LLM-based predictor outperforms fully supervised GNNs even using only a \texttt{text-only prompt}.
Therefore, we first explore the ability of a single LLM to perform predictive tasks in DyTAGs. 
Following the attempts in static TAGs~\cite{LLM_explor_Zhikai_23}, we study the effect of \texttt{text-only prompts} and \texttt{structure-aware} prompts for LLM-based predictors separately. \\
\textbf{Text-only prompt.}
In the \texttt{text\_prompt}, only node text is used for prediction.
In the \texttt{text\_few\_shot\_prompt}, we additionally include six samples of question-answer pairs constructed from the validation set in the prompt.~\footnote{In this paper, for the collection of examples, we only collect one sample per batch to ensure sample diversity. 
This is because the samples within the same batch are temporally close, often leading to repeated nodes and pairs.} \\
\textbf{Structure-aware prompt.}
In the DyTAGs task, constructing effective \texttt{structure\_aware\_prompts} faces several challenges.
For example, in LP, where the objective is to predict whether $(u, v)$ will interact in the future, incorporating complete historical interaction data for both nodes leads to excessively long prompt lengths. 
For a node with 100 historical interactions, even if each interaction’s information can be recorded within 100 tokens, it would require 10,000 tokens for each prediction, not to mention information beyond 1-hop.
In static TAGs, a common approach is to sample over the neighbors of $u$ and $v$, and only include the sampled neighbors in the prompts. 
However, in DyTAGs, this approach becomes ineffective, as most of $u$'s historical interactions are not directly related to $v$.
Therefore, a more effective $g(\cdot)$ in Equation~\ref{eq:1} is needed to extract the relevant structural information for pair $(u, v)$.

Fortunately, recent works have shown that using heuristic structural metrics can yield good performance.
For instance, EdgeBank, which only uses the existence of historical interactions, has been validated as a strong baseline in LP~\cite{Edgebank_22}. 
In~\cite{TGB_23}, heuristic methods outperform GNNs in the future node property prediction task (similar to EC but node-wise).
These findings motivate us to extract structural metrics rather than exhaustive historical interactions during prediction.
Specifically, at time $t_c$ we extract the following structural metrics as $M_{t_c}$ for pair $(u,v)$:
\begin{itemize}[leftmargin=*]
    \item The Historical Interaction (HI) count between nodes \(u\) and \(v\), which is defined as:
    \(
    \text{HI} = \left| { (u, v, t) \in \mathcal{E}, t<t_c } \right|.
    \)
    \item The number of Common Neighbors (CN) between \(u\) and \(v\), which is defined as:
    \(
    \text{CN} = \left| \mathcal{N}_{t_c}(u) \cap \mathcal{N}_{t_c}(v) \right|.
    \)
    \item The node frequency (NF) of $u$ and $v$, respectively, as well as the average interaction frequencies of their neighbors.
    Note that globally, due to the setting of DTGB~\cite{DTGB_2024} where positive and negative samples share the same source nodes, only the destination node frequency (DNF) exhibits a global distinction.  
    \item The historical Edge Label Distribution (ELD) of \((u, v)\), including $p_t(u), p_t(v)$, and $p_t(u,v)$.
\end{itemize} 
\textbf{Knowledge Definition.} 
After extracting metrics as $M_{t_c}$, we define the corresponding knowledge $K$ following the inductive bias of previous works~\cite{Edgebank_22, DygFormer_23, TGB_23}.
In task LP, we assume the higher the HI and CN, the more likely it is to form interactions between $u$ and $v$ in the future.
Additionally, if the destination node's frequency correlates with the source node's preferences, e.g., a high average neighbor frequency of the source and a high frequency for the destination, future interactions are more likely.
In task EC, we assume that edge labels tend to remain consistent over time. 
Specifically, the more frequently a label appears in the historical data, the more likely it is to appear in the future.
In the NR task, the knowledge is similar to that in LP, but the large number of candidate destination nodes requires the pre-filtering of negative samples.
To achieve this, we first recall pairs \( (u,v) \) that satisfy \( \text{HI} > 0 \) or \( \text{CN} > 0 \), as these metrics are associated with positive samples when high by the knowledge.
The destination nodes in the recalled pairs form the final candidate set.
We require the predictor to output the likelihood (\( \in [0, 1] \)) for each candidate in the final candidate set and calculate the rank of the positive sample based on its likelihood value.

\begin{figure}[htbp!]
\centering\includegraphics[width=\linewidth]{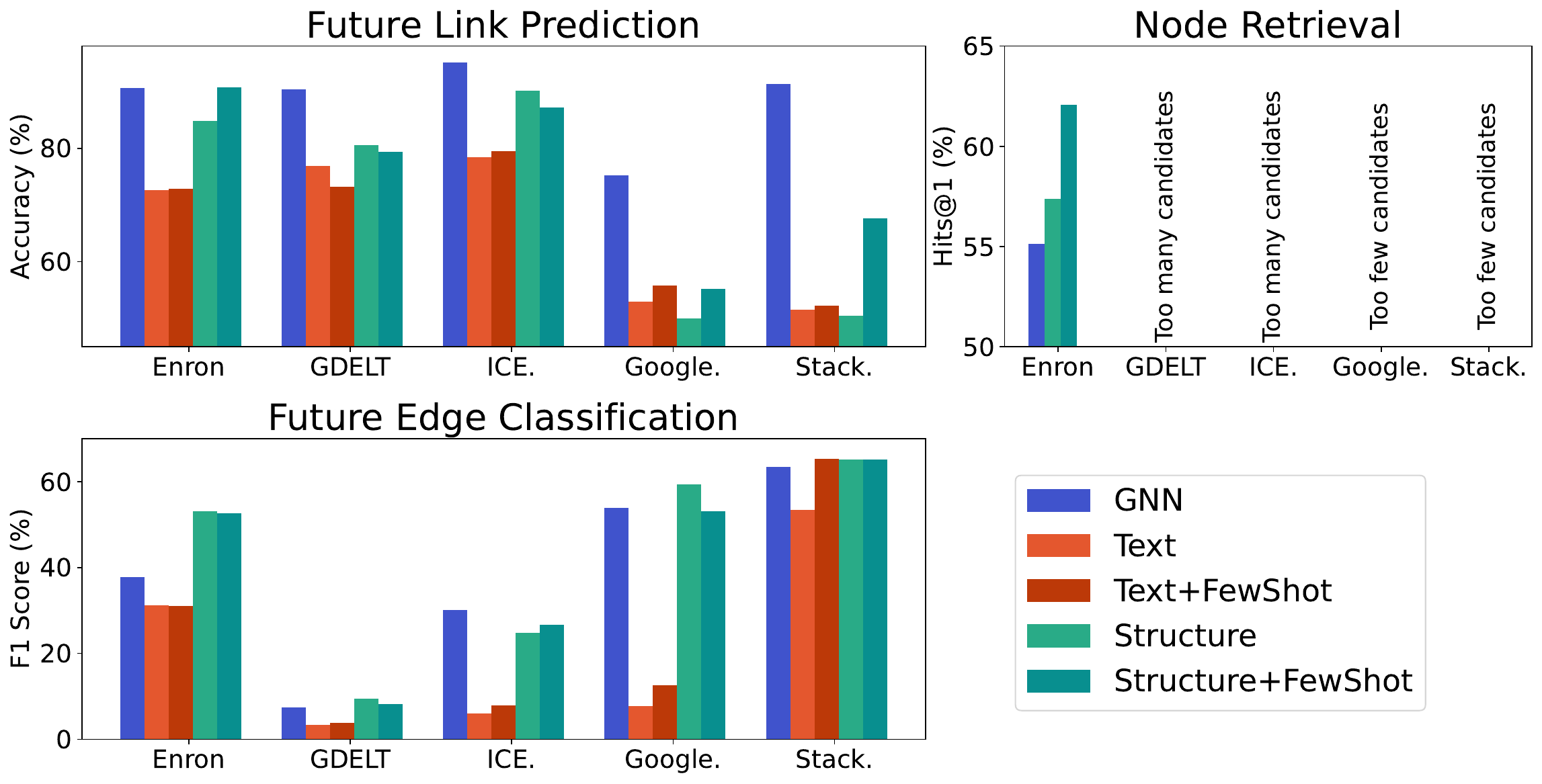}
    \caption{Performance comparison between GNNs (average over TCL, GraphMixer, DygFormer) and LLM-based predictors (average over GPT and DeepSeek-based backbones).}
    \label{fig:compare}
\end{figure}

As time evolves, $M_{t_c}$ is dynamically updated, while $K$ remains fixed.
In the prompts to the predictor LLM, both $M_{t_c}$ and $K$ are included together with task descriptions. 
The experimental results are presented in Figure~\ref{fig:compare}.

\Obs \textbf{Structural information significantly enhances the performance of LLM-based predictors when relevant.} 
From the results, we observe that \texttt{text-only} prompt performs significantly worse than \texttt{structure-aware prompt} in most cases, indicating that in DyTAGs, the role of textual information is less significant compared to static TAGs.
In the future edge classification task, structure-aware LLMs outperform supervised GNNs.
In the link prediction task, in datasets such as Enron, GDELT, and ICEWS1819, structure-aware LLMs achieve performance comparable to that of GNNs.
The results demonstrate that when the knowledge and metrics are effective, LLM-based predictors hold significant potential.
\Obs \textbf{Desirable knowledge significantly varies between different domains.} 
Despite the progress in email graphs and knowledge graphs, a noticeable performance gap emerges when \texttt{structure-aware} prompt is applied to Googlemap\_CT and Stack\_elec, where their performance significantly lags behind GNNs and even \texttt{text-only} prompts.
While few-shot examples could help the LLMs sometimes, the performance gain is inconsistent. 

The performance degradation is primarily due to the differing applicability of the knowledge across domains. 
For instance, common neighbors are strongly helpful in LP in non-bipartite graphs, as a higher number of common neighbors indicates a significantly higher likelihood of positive samples.  
However, in bipartite graphs such as GoogleMap\_CT and Stack\_Elec, both positive and negative samples have zero common neighbors, rendering this metric irrelevant.  
As a result, misapplying this knowledge causes the LLM to misclassify positives as negatives in both datasets, yielding nearly 50\% accuracy, which is even worse than \texttt{text-only} prompts.  

Beyond improper knowledge usage, knowledge can naturally contradict across datasets.  
For example, in GoogleMap\_CT, positive samples have higher DNF than negatives.  
In the validation set, over 93\% of positive pairs exhibit DNF values greater than 5, compared to only 77\% of negative pairs.  
Conversely, in Stack\_Elec, 78\% of positive samples have zero DNF, compared to only 27\% of negative pairs, indicating a preference for new destination nodes.  

Domain-specific knowledge differences further hinder the node retrieval task. 
During the experiments, the single LLM only works in the Enron dataset. 
The filtering process recalls too many samples on GDELT and ICEWS1819, resulting in excessively long input contexts.
It also filters out all samples on GoogleMap\_CT and Stack\_Elec, leading to poor retrieval performance. 
The above failure suggests that the recall strategy is influenced by the variation in desirable knowledge across domains. 
Consequently, node retrieval remains a challenging task for a single LLM-based predictor.

\section{The GAD Framework}
\label{sec:DAG}
From the results in Section~\ref{Sec:given text}, we can summarize several drawbacks of employing a single LLM as the predictor due to the characteristics of dynamic graphs:  
\textbf{C1: Dataset-specific knowledge requirements.} Different domains require distinct knowledge.
    When the knowledge is misaligned, the performance significantly degrades, and failures occur in the recall phase of the node retrieval task.  
\textbf{C2: Lack of node-specific adaptation.} The knowledge is the same for all nodes without capturing fine-grained node-wise differences.  
    The extracted information is often too long for the predictor LLM to process effectively.  
\textbf{C3: Lack of knowledge update mechanism.} Knowledge remains static after initialization. In a temporally evolving dynamic graph system, fixed knowledge is prone to failure as it may shift or become biased.

To address the above challenges, we propose the GraphAgent-Dynamic (GAD) framework.
Figure~\ref{fig:framework} illustrates the role specialization, workflow, and structured collaboration within this framework. 
The core idea is to decompose the prediction task into multiple sub-tasks, each handled by a separate LLM-empowered agent.  
To address \textbf{C1}, we introduce global summary agents \( f_g \) to generate dataset-specific knowledge.  
To address \textbf{C2}, we incorporate local summary agents \( f_l \) to summarize node-wise profiles that capture fine-grained knowledge.  
To address \textbf{C3}, we include knowledge reflection agents \( f_r \) that update the obtained knowledge.  
The overall workflow of GAD is formulated as:  
\begin{align*}  
M_t &= g(\mathcal{G}, t) \quad &\text{(Information Extraction)} \\ 
K_g &= f_g(M_t), \quad K_l = f_l(M_t) \quad &\text{(Knowledge Generation)} \\  
P_{\text{tmp}} &= f(M_t, K_g, K_l) \quad &\text{(Surrogate Prediction)} \\  
K_{\text{new}} &= f_r(K_g, K_l, P_{\text{tmp}}, M_t) \quad &\text{(Knowledge Reflection)} \\  
\quad P &= f(M_t, K_g, K_l, K_{\text{new}}) \quad &\text{(Final Prediction)}  
\end{align*}
In the following sections, we provide a detailed explanation of the design of each agent.  

\subsection{Global Summary Agent $f_g$}
Global Summary Agents aim to extract dataset-specific knowledge.  
The setup of Global Summary Agents consists of the following steps:  \\
\textbf{Data Preparation.}  
We extract HI, CN, DNF, and ELD metrics from the validation set.  
For metrics HI, CN, and DNF, we compute their distributions for both positive and negative samples.  
Specifically, for each metric \( m \), we define its distribution as a dictionary \( \{ \text{key}: \text{value} \} \), where the key represents a condition (e.g., \( HI = k \)) and the value denotes the proportion of samples satisfying that condition.  
Formally, the dictionary is defined as:  
\[
\mathcal{D}_m = \{m=0: \mathbb{P}(m=0), m>0: \mathbb{P}(m>0), \dots, m>5: \mathbb{P}(m>5)\}.
\]
After generating separate dictionaries for positive and negative samples, both are used together as input to the agents.  

For the ELD metric, we define a frequency-preference dictionary, initialized as \( \{c_1': 0, c_2': 0, c_3': 0, c'_{\text{others}}: 0 \} \), where the keys represent the preferences for the top-1, top-2, top-3 historical labels, or other labels. 
We categorize each sample into one of these classes and enumerate the validation set to complete the dictionary.  
For instance, consider a pair \( (u, v, t) \) that has been historically labeled \( \{ c_1, c_2, c_3 \} \), ordered by descending frequency. 
If its label at time \( t \) is \( c_1 \), we increment \( c_1' \) by 1, indicating that the most frequent historical label is chosen. 
We collect three distinct preference dictionaries for node \( u \), node \( v \), and the pair \( (u, v) \), respectively.
After processing all samples, the values in each ELD dictionary are normalized to yield percentage values.

For textual information, we collect node and edge text data by sampling 30 text instances at uniform time intervals from the validation data. 
The text is truncated to a length of 50 to ensure the input does not exceed the maximum context length. \\
\textbf{Knowledge Generation.} LLMs exhibit emergent step-by-step reasoning capabilities, enabling them to iteratively approach the final answer by decomposing tasks into sub-problems~\cite{sub_task}. 
Building on this insight, we decompose the knowledge generation process into sub-tasks, each handled by a separate agent. 
Specifically, we deploy two groups of agents: one for structural prediction and one for edge label classification. 
Each group comprises two agents: a structural summary agent and a text summary agent.
The text summary agent focuses on summarizing how to utilize text to solve tasks, while the structural summary agent focuses on utilizing the structural metrics for its corresponding task.
For each of these four agents, we require the generation of the following components: 
\begin{itemize}[leftmargin=*]
    \item \textbf{Metric Significance:} This component assigns a significance rank to each metric based on its relevance to downstream tasks, including ``Extremely Significant'', ``Helpful'', ``Maybe Related'' and ``Not Relevant''.
    The input to the predictor excludes metrics that fail to demonstrate substantial relevance.
    \item \textbf{Knowledge:} This component provides a rationale for the significance ranking of each metric, together with practical guidelines for its application. 
    This step aims to establish clear knowledge for identifying positive samples.
    \item \textbf{Threshold (Optional):} For numerical metrics \( \text{HI} \), \( \text{CN} \), and \( \text{DNF} \), decision boundaries (positive/negative sample thresholds) are derived based on the distributional analysis. 
    For example, in Enron, the threshold is determined as $\text{HI} < 1$ and $\text{CN} < 1$ by the agent.
    When the threshold is met, we can directly determine a sample as negative and exclude it from the final candidate set.
\end{itemize}

\subsection{Local Summary Agent $f_l$}
To generate node-specific knowledge, we employ Local Summary Agents to generate node-wise profiles to capture node-specific preference.
The agent initialization and data preparation processes are similar to those in the global knowledge summary. 
We initialize structural and text agents, respectively. 
During the data preparation phase, for each node, we extract metrics HI, CN, NF, and ELD, as well as the related node text, neighbors' text, and edge labels. 
Subsequently, we generate the following profile for each node:
\begin{itemize}[leftmargin=*]
    \item \textbf{Node Description:} A description of the node itself.
    \item \textbf{Neighbor Preference:} A summary of the node's neighbor preference based on the text of its historical neighbors.
    \item \textbf{Edge Label Preference:} A summary of the node's preferences for specific edge labels based on edge label text.
    \item \textbf{Structural Preference:} The node's structural preferences on metrics. 
    Given the limited local data, this preference is likely to mislead global knowledge. 
    Thus, if structural preferences are not significant, we encourage the agent to mark them as ``Not Significant'' and forbid their use in the future.
\end{itemize}

Since DyTAGs are large in size and sparse in interactions (i.e., most nodes participate in very few interactions), to reduce computational complexity, we generate profiles for the top 10\% of the most active nodes, following the Pareto Principle. 
In Appendix~\ref{app:pareto}, we show the contribution of these nodes to the overall interactions. 
It can be observed that the top 10\% of nodes contribute to more than 70\% of the interactions.



\subsection{Knowledge Reflection Agent $f_r$}
In the global knowledge generation phase, we encourage agents to mine metrics that identify positive samples actively. 
However, existing knowledge may produce false positives when the metrics are considered independently.
For instance, in GDELT, both high HI and CN generally favor positive samples. 
However, when CN is high, and HI is low, creating contradictory knowledge between the two metrics, the likelihood of the sample being negative increases significantly. 
Despite this, the agent may still output a positive prediction due to its incomplete knowledge.
To address this issue, we develop a reflection agent that adaptively learns from prediction accuracy feedback across different graph domains.
It includes the two steps below:
\begin{itemize}[leftmargin=*]
    \item \textbf{Prediction Trajectory.}
    To reinforce the reflective agent, we first collect ``verbal'' feedback from previous experiences during prediction. 
    Specifically, we employ a temporary predictor agent to make predictions on the validation set. 
    Our focus is on false positive errors, as the limited metrics available for identifying positive samples during the global knowledge phase encourage the agent to utilize metrics actively. 
    In this phase, false positive errors are more likely to occur, while false negative errors primarily arise from insufficient metric discovery. 
    We record 50 prediction trajectories for negative samples, including the task instructions, input knowledge, and output prediction accuracy as the input to the agent.

    \item \textbf{Self-Reflection.} Based on the trajectory, we require the reflection agent to generate: 1) \textbf{Significance.} If the predictor already has high classification accuracy, it indicates that the knowledge quality is desirable and does not require supplementation. 
    On the other hand, if supplementation significantly contradicts existing knowledge, it may introduce erroneous knowledge due to overly aggressive updates. Both cases suggest a failure in reflection and are marked as ``Not Significant''.
    2) \textbf{Supplementation.} 
    The supplementation of existing global knowledge $K_{\text{new}}$ that helps reduce false positive errors is generated.
    The generated supplementation, combined with the global knowledge, is provided to the predictor agent.
\end{itemize}

\subsection{The Predictor Agent}
The design of the predictor follows Equation~\ref{eq:1}, with the modification that \( K \) is replaced by $K_g$, \( K_l \) and \( K_{\text{new}} \).
The guidance in $K_g$ provided by the Global Summary Agents is included, but only for metrics marked as ``Extremely Significant'' or ``Helpful''. 
If no such metrics are available, the ``Maybe Related'' metrics are utilized. 
``Not Relevant'' metrics are excluded to reduce hallucination. 
Supplementation $K_{\text{new}}$ from the Knowledge Reflection Agent is combined with $K_g$ if marked as ``Significant''.
Additionally, if the node involved in each prediction has a local summary, it is included as local knowledge $K_l$.

For node retrieval, only samples that satisfy the threshold in $K_g$ are included in the final candidate set. 
The candidate set is then sorted according to the preference in $K_g$. 
For example, if high HI is favored, candidates are sorted in descending order based on their HI values. 
In cases where different metrics contradict, priority is given based on their significance level. If the candidate set exceeds 20 samples, we only retain the top 20 as the final candidates. 
Only the relevant information of final candidates is included in the prompt to the Predictor Agent.

During evaluation, $K_g$, $K_{\text{new}}$ and $K_{l}$ remain constant, while the information $M_t$ is updated over time, reflecting the evolving observations. 
The full prompts are provided in Appendix~\ref{app:prompts}, and generated thresholds and other output examples are provided in Appendix~\ref{app:examples}.

\section{Experiments}\label{sec:exp}

\subsection{Performance of GAD}

\begin{table*}[ht]
\centering
\caption{Performance comparison of GNNs and single LLM models for Link Prediction (LP), Edge Classification (EC), and Node Retrieval (NR) tasks. 
We report Accuracy (\%) for the LP task, 
F1 score (\%) for the EC and hits@1 (\%) for NR tasks. 
The best performance for GNNs is highlighted in {\cred{red}}, and the best LLM-based predictors in {\cblue{blue}}.}
\label{tab:full}
\resizebox{\textwidth}{!}{%
\begin{tabular}{ll|rrr|rrrr|rrrr|rr}
\toprule
\multirow{2}{*}{\textbf{Task}} & \multirow{2}{*}{\textbf{Dataset}} & 
\multicolumn{3}{c|}{\textbf{GNNs}} & 
\multicolumn{2}{c}{\textbf{Text}} & \multicolumn{2}{c|}{\textbf{Text-FewShot}} & \multicolumn{2}{c}{\textbf{Structure}} & \multicolumn{2}{c|}{\textbf{Structure-FewShot}} & \multicolumn{2}{c}{\textbf{GAD}} \\
\cmidrule(lr){3-5} \cmidrule(lr){6-7} \cmidrule(lr){8-9} \cmidrule(lr){10-11} \cmidrule(lr){12-13} \cmidrule(lr){14-15}
 & & TCL & GraphMixer & DygFormer & GPT & DeepSeek & GPT & DeepSeek & GPT & DeepSeek & GPT & DeepSeek & GPT & DeepSeek \\
\midrule
\multirow{5}{*}{\textbf{LP}} & Enron & 90.56 & 88.08 & \cred{93.18} & 72.59 & 67.28 & 72.90 & 72.85 & 84.80 & 83.37 & 90.78 & 93.81 & 94.09 & \cblue{94.72}\\
 & GDELT & 90.75 & 89.43 & \cred{91.10} & 76.88 & 72.68 & 73.25 & 72.17 & 80.61 & 86.11 & 79.33 & 83.39 & 77.53 & \cblue{86.20}\\
 & ICEWS1819 & \cred{95.92} & 94.30 & 95.23 & 78.37 & 83.88 & 79.52 & 85.95 & 90.13 & 91.35 & 87.23 & \cblue{93.00} & 89.31 & 92.16 \\
 & Google. & \cred{77.51} & 74.00 & 74.23 & 53.02 & 55.15 & 55.83 & 59.33 & 50.00 & 50.10 & 55.17 & 52.41 & \cblue{63.31} & 60.17\\
 & Stack\_elec & 91.24 & 91.19 & \cred{91.48} & 51.57 & 51.46 & 52.20 & 52.61 & 50.47 & 51.34 & 67.66 & 57.03 & 55.63 & \cblue{70.16} \\
\midrule
\multirow{5}{*}{\textbf{EC}} & Enron & 18.40 & \cred{48.36} & 46.75 & 31.22 & 28.02 & 31.12 & 43.18 & 53.03 & 53.93 & 52.65 & \cblue{53.97} & 53.74 & 53.87\\
 & GDELT & 1.26 & 8.92 & \cred{12.15} & 3.44 & 4.44 & 3.86 & 5.54 & 9.44 & \cblue{12.89} & 8.25 & 11.62 & 10.42 & \cblue{13.16}\\
 & ICEWS1819 & 29.66 & 29.24 & \cred{31.30} & 6.09 & 4.85 & 7.83 & 10.92 & 24.82 & 29.77 & 26.65 & \cblue{30.21} & 26.41 & 28.59\\
 & Google. & 51.09 & 55.17 & \cred{55.18} & 7.69 & 4.16 & 12.53 & 37.11 & 59.42 & \cblue{59.78} & 53.11 & 57.64 & 55.96 & 55.96\\
 & Stack\_elec & 62.54 & 62.57 & \cred{65.24} & 53.47 & 65.04 & 65.35 & 65.51 & 65.13 & 64.90 & \cblue{68.03} & 64.16 & 65.48 & 62.98\\
\midrule
\multirow{5}{*}{\textbf{NR}} & Enron & 46.46 & 42.90 & \cred{76.04} & - & - & - & - & 57.38 & 64.01 & 62.06 & 65.40 & 
\cblue{72.79} & 72.21\\
 & GDELT & 44.53 & 39.40 & \cred{46.25} & - & - & - & - & - & -& - & - & 50.01 & \cblue{50.03} \\
 & ICEWS1819 & \cred{81.32} & 74.48 & 80.99 & - & - & - & - & - & - & - & -& 76.18 & \cblue{76.52} \\
 & Google. & \cred{15.84} & 10.83 & 13.86 & - & - & - & - & - & - & - & - & 8.85 & \cblue{10.27} \\
 & Stack\_elec & 7.63 & \cred{26.00} & 8.22 & - & - & - & - & - & - & - & - & 3.58 & \cblue{4.82} \\
\bottomrule
\end{tabular}%
}
\end{table*}
\Obs \textbf{GAD demonstrates improved generalization to tasks with diverse, context-dependent rules across datasets.}
The full experiment results are shown in Table~\ref{tab:full}.
The results show that GAD outperforms the single predictor on LP tasks, particularly on GoogleMap\_CT and Stack\_Elec, where domain knowledge shifts.  
By leveraging knowledge from Global Summary Agents, the predictor correctly identifies node frequency as the key metric and accurately differentiates its usage across the two datasets.  
Moreover, through multi-agent collaboration, LLM-based predictors successfully perform node retrieval across all datasets, demonstrating the superiority of GAD over single agents.  

\begin{table}[ht]
    \centering
    \caption{Ablation studies of key components of GAD using GPT4o-mini as the backbone.}
    \label{tab:ab}
    \resizebox{\linewidth}{!}{%
    \begin{tabular}{l|cccccc}
    \toprule
    & Enron & GDELT & ICEWS1819 & Google. & Stack\_elec \\
    \midrule
    GAD & 94.09 & 77.53 & 89.95 & 63.31 & 55.63 \\
    w/o Local & 93.71 & 76.64 & 89.31 & 60.47 & 50.62 \\
    w/o Reflection & -  & 69.55 & 87.47 & 59.78 & - \\
    \bottomrule
    \end{tabular}
    }
    \vspace{-0.5em}
\end{table}

\subsection{Ablation Study}
In the ablation study, we validate the effectiveness of Local Summary Agents and Knowledge Reflection Agents. 
The results of the ablation study of GAD are shown in Table~\ref{tab:ab}.
The experimental results demonstrate the efficacy of both modules in enhancing performance. 
In practical applications, Local Summary Agents can provide better support for personalized tasks, while Knowledge Reflection Agents offer a sustainable update mechanism to adapt to long-term tasks. 
The effectiveness of these modules ensures the integrity of the GAD Framework.

\subsection{Limitations of GAD}
\Obs\textbf{GAD benefits little over a single LLM-based predictor in scenarios with unified rules.}
In the future edge classification task, although GAD generally outperforms GNNs, it does not show a clear advantage over a single LLM as the predictor.  
Upon examining the outputs, we observe that the Global Summary Agent consistently ranks Edge Text as more important than ELD, assigning Node Text only ``Maybe Related'' across all datasets.  
Since Edge Text is forbidden to be used, the remaining guidance from the global summary remains a subset of the knowledge outlined in Section~\ref{Sec:given text}, only suggesting that labels with higher historical frequency are more likely to appear in the future.
Similarly, in datasets like Enron, GDELT, and ICEWS1819, where human-written knowledge effectively guides the single LLM in link prediction, GAD provides only marginal performance gains.  

So, the primary strength of the GAD framework lies in its adaptability to tasks with diverse, task-specific knowledge. 
In cases where tasks and rules are consistent across all scenarios, good human-written instruction can lead to comparable or better performance.    
In the next section, we will explore the potential of enhancing the performance of LLM-based predictors on specific tasks.  
\section{Further Improvement}
As a unified framework for various downstream tasks, GAD generally outperforms a single predictor across diverse domains without any dataset-specific training.
However, it is natural to trade off generalizability for performance by incorporating dataset-specific knowledge. 
With substantial human effort in data mining, this approach can yield superior results on particular datasets or tasks. 
In this section, we explore strategies for further enhancing LLM-based methods tailored to specific datasets and tasks.
\subsection{Dataset-Specific Fine-tuning for LP}
If we relax the constraints on LLM-based predictors and allow dataset-specific tuning, we can improve performance on specific datasets.  
Specifically, we extract few-shot examples and fine-tune the LLMs' parameters using supervised fine-tuning (SFT).
We apply Qlora~\cite{qlora} with 4-bit quantization.
We use Llama 3~\cite{llama3modelcard} as the backbone with \texttt{structure-aware} prompt, and propose the following three fine-tuning variants:
\begin{itemize}[leftmargin=*]
    \item \textbf{SFT}: For each dataset, we extract 10,000 question-and-answer pairs from the validation set for each task (30,000 pairs in total) and use these to tune the model. 
    During the evaluation of a specific dataset, we use the model tuned on it to make predictions.
    \item \textbf{SFT-All}: For all datasets, we extract 5,000 question-and-answer pairs per task from each dataset (75,000 pairs in total) and tune a unified model that handles all prediction tasks.
    \item \textbf{SFT-\{D\}}: To test the transferability of models, we use the Llama-specific model tuned on dataset \{D\}, denoted as SFT-\{D\}, to make predictions in other datasets. 
    Specifically, we use Enron and Googlemap\_CT\ to evaluate the transferability across datasets.
\end{itemize}

\begin{table}[ht]
\centering
\vspace{-0.5em}
\caption{Performance of SFT predictors in LP and EC.
The best performance is highlight in \cred{red}.}
\label{tab:sft}
\resizebox{\linewidth}{!}{%
\begin{tabular}{ll|rr|rrrr}
\toprule
\textbf{Task} & \textbf{Dataset} & 
\textbf{GNN} & \textbf{GAD}
& \textbf{SFT} 
& \textbf{SFT-All}
& \textbf{SFT-Enron}
& \textbf{SFT-Google.} \\
\midrule
\multirow{5}{*}{\textbf{LP}} & Enron & 93.18 & 94.72 & 94.40 & \cred{95.36} & - & 72.58 \\
& GDELT & 91.10 & 86.20 & \cred{91.95} & 90.94 & 86.77 & 87.62 \\
& ICEWS1819 & \cred{95.92} & 92.16 & 95.47 & 95.34 & 92.93 & 86.78\\
& Google. & \cred{77.51} & 63.31 & 69.79 & 69.68 & 49.99 & - \\
& Stack\_elec & \cred{91.48} & 70.16 & 82.50 & 76.71 & 71.30 & 56.42 \\
\midrule
\multirow{5}{*}{\textbf{EC}} & Enron & 48.36 & 53.87 & 54.30 & 54.16 & - & \cred{55.01}\\
 & GDELT & 12.15 & \cred{13.16} & 4.72 & 7.75 & 5.73 & 5.06 \\
 & ICEWS1819 & \cred{31.30} & 28.59 & 14.22 & 7.75 & 15.82 & 14.40 \\
 & Google. & 55.18 & 55.96 & 57.34 & \cred{60.46} & 54.75 & - \\
 & Stack\_elec & 65.24 & 65.09 & \cred{68.07} & 68.04 & 65.41 & 67.05 \\
\bottomrule
\end{tabular}%
}
\end{table}
After tuning, we use these models as predictors. 
The results are shown in Table~\ref{tab:sft}.

\Obs\textbf{SFT significantly enhances performance in LP.}
The results for the LP task show that SFT notably improves the performance of LLM-based models. 
After being tuned, a single LLM-based predictor surpasses the best-performing GNN variant on 2 out of 5 datasets, indicating that SFT can better capture fine-grained semantic patterns within metrics and knowledge.

\Obs\textbf{Transferability is the main challenge of SFT.}  
In future link prediction, domain-specific SFT tuning achieves optimal performance but suffers significant degradation when applied to dissimilar domains.  
SFT-All maintains comparable performance across datasets, but its performance deteriorates compared to dataset-specific tuning due to the mixing of knowledge from different domains.  
Worse yet, when tuned solely on GoogleMap\_CT, the predictor becomes significantly less effective on datasets from other domains.  
These results highlight that the lack of cross-domain adaptability remains a major challenge for single LLM-based predictors.  
Consequently, multi-agent approaches are still necessary to ensure robust performance across diverse tasks.  

\subsection{Better Recallers Help For NR}
In Table~\ref{tab:node_retrieval} in the Appendix, we present the complete results including Hits@1, Hits@3, and Hits@10 for NR.  
We observe that the largest gap between GAD and GNN occurs at Hits@10, indicating that the main limitation of LLM-based predictors lies in recalling the final candidate nodes.  
We currently employ a rule-based approach for the recall step to ensure the workflow remains LLM-based while maintaining inference efficiency.  
However, the accuracy of recall is challenging to guarantee.  
A more effective recall design could further improve the performance of LLM-based predictors.  

In Table~\ref{tab:NR_GNN}, we explore the potential of a dataset-specific trained recaller. 
We use the best GNN in LP in each dataset to recall the top-10 candidate destination nodes, followed by a GAD for the final retrieval. 
As shown in the results, better recall accuracy leads to improvements across most datasets.
\begin{table}[h]
    \centering    
    \caption{Performance comparison of GAD and GAD with GNN as the recaller in node retrieval.}
    \label{tab:NR_GNN}
    \resizebox{\linewidth}{!}{
    \begin{tabular}{llccccc}
        \toprule
        & Method & Enron & GDELT & ICEWS1819 & Google & Stack\_elec \\
        \midrule
        \multirow{2}{*}{Hits@1} 
        & GAD     & 72.79 & 50.01 & 76.18 & 8.85  & 3.58 \\
        & GAD+GNN & 73.19 & 49.23 & 78.03 & 13.34 & 8.37 \\
        \midrule
        \multirow{2}{*}{Hits@3} 
        & GAD     & 84.28 & 71.90 & 87.52 & 20.69 & 13.44 \\
        & GAD+GNN & 85.65 & 73.11 & 90.51 & 27.21 & 22.16 \\
        \bottomrule
    \end{tabular}
    }
    \vspace{-0.5em}
\end{table}

\subsection{Intrinsic Challenges in EC}
Despite the success of dataset-specific strategy in improving performance in LP and NR, we see that SFT does not lead to significant or consistent performance improvements.  
Given that neither SFT nor GAD have further enhanced the performance of LLM-based predictors on this task, we reflect on the task design to explore potential improvements.  \\
\textbf{Intrinsic limits for EC.}
Future edge classification is inherently a single-label classification task.  
However, if a node pair with the same attributes receives different edge labels over time, it implies a multi-label setting, imposing an intrinsic performance upper bound since only one label can be output.  
We define \textbf{Label Consistency} to measure the extent to which a task on a dataset tends toward multi-label classification:  
\[
\text{Label Consistency} = \frac{
  \sum_{(u,v), \text{count}(u,v)>1} 
  \mathbf{1}(\text{label}(u,v) = \hat{\text{label}}_{(u,v)})
}{
  \sum_{(u,v), \text{count}(u,v)>1}
},
\]  
where $\hat{\text{label}}_{(u,v)}$ represents the most frequent edge label for pair $(u, v)$ of the same attributes.  
In \textbf{Label Consistency}, we examine pairs with identical source and destination nodes (i.e., same node text and history).  
In \textbf{Text Consistency}, we focus on pairs with identical edge text.  
A higher value reflects better label consistency given the same attributes.
Some statistics related to GoogleMap\_CT and Stack\_elec are omitted due to the scarcity of pairs that appear more than once (less than 10\% of the total pairs).  
\begin{table}[h]
    \centering
    \caption{The statistics of label consistency (\%) in the future edge classification task. 
    }
    \label{tab:ec_ana}
    \resizebox{\linewidth}{!}{
    \begin{tabular}{cccccc}
        \toprule
         & Enron & GDELT & ICEWS1819 & Google. & Stack\_elec \\ 
         \midrule
        Pair Consistency & 64.9 & 29.4 & 48.5 & - & 94.5\\
        Text Consistency & 68.6 & 100 & 100 & 77.3 & -\\
        \bottomrule
    \end{tabular}
    }
\end{table}
In Table~\ref{tab:ec_ana}, we present the calculated consistency, which shows that Pair Consistency is notably low in datasets like GDELT and ICEWS1819, indicating that the same pairs frequently receive different edge labels.  
Since only one label can be assigned per edge, this naturally results in over 50\% of samples being misclassified.  
Interestingly, the high text consistency in these datasets suggests that edge labels are primarily determined by the associated edge text.  
Thus, these datasets have little room for further improvement when edge text is excluded. \\
\textbf{Edge classification with edge text.}  
Since edge text can be decisive in determining edge labels, we explore a new setting where edge text from the interaction \((u, v)\) is incorporated to determine its class.  
For GNNs, we include edge embeddings in the edge classifier. 
For LLM-based predictors, we include raw edge text in the prompt.  
Interestingly, introducing edge text does not always simplify the task.  
In Enron and GoogleMap\_CT, we observe low text consistency, indicating that identical edge text can correspond to different edge labels.  
Table~\ref{tab:ec_with_text} presents the results of edge classification in these two datasets.  
Compared to Table~\ref{tab:full}, the performance of a single LLM-based predictor degrades in Enron, while GAD consistently holds superiority.  
This suggests that edge classification remains a challenging yet compelling task in certain datasets, and GAD retains its advantage in scenarios requiring diverse knowledge.  

\begin{table}[ht]
    \centering
    \caption{The weighted F1 scores (\%) of GNNs and LLM-based predictors in edge classification, where edge text is utilized. 
    LLM-based predictors employ GPT4o-mini as backbones.}
    \label{tab:ec_with_text}
    \resizebox{\linewidth}{!}{
    \begin{tabular}{cccccc}
        \toprule
         & TCL & GraphMixer & DygFormer & Structure & GAD\\ 
         \midrule
         Enron & 39.72 & 48.78 & 50.83 & 48.38 & \cred{53.11} \\
         Google & 64.52 & 67.16 & 67.12 & 68.86 & \cred{69.47} \\
        \bottomrule
    \end{tabular} 
    }
    \vspace{-0.5em}
\end{table}

\section{Conclusion}
In this paper, we investigate the ability of LLMs-as-predictors to address prediction tasks on DyTAG.
We find that, compared to static graphs, structural information is more important on dynamic graphs, domain differences become more pronounced, and task formats pose greater challenges for single-LLM-based predictors. 
To adapt to diverse scenarios and multi-task settings, we propose a multi-LLM-based agent collaboration framework, GAD, which exhibits performance comparable to GNNs and even surpasses them on several tasks, all without requiring any training.
This framework better aligns domain knowledge and is more suitable for long-term, domain-shifting prediction tasks. 
Finally, we explore potential avenues for further improving LLM-based predictors. 
We identify a series of challenges in designing LLM-based predictors for DyTAG, offering insights for the future development of Dynamic Graph Foundation Models.

\begin{acks}
This research was supported in part by National Natural Science Foundation of China (No. 92470128, No. U2241212), by National Science and Technology Major Project (2022ZD0114802), by Beijing Outstanding Young Scientist Program No.BJJWZYJH012019100020098, by the National Key Research and Development Plan of China (2023YFB4502305) and Ant Group Research Fund. 
We also wish to acknowledge the support provided by the fund for building world-class universities (disciplines) of Renmin University of China, by Engineering Research Center of Next-Generation Intelligent Search and Recommendation, Ministry of Education, by Intelligent Social Governance Interdisciplinary Platform, Major Innovation \& Planning Interdisciplinary Platform for the “Double-First Class” Initiative, Public Policy and Decision-making Research Lab, and Public Computing Cloud, Renmin University of China.
The work was partially done at Gaoling School of Artificial Intelligence, Beijing Key Laboratory of Big Data Management and Analysis Methods, MOE Key Lab of Data Engineering and Knowledge Engineering, and Pazhou Laboratory (Huangpu), Guangzhou, Guangdong 510555, China.
\end{acks}

\bibliographystyle{ACM-Reference-Format}
\bibliography{reference}

\appendix
\section{Dataset Statistics} \label{app:data_stats}
The detailed statistics of the datasets are provided in Table~\ref{app:data_stats}.
For a detailed dataset description, we refer to Section B.1 in the Appendix in DTGB~\cite{DTGB_2024}.
These dataset descriptions are also fed to the Initial Agent to extract dataset properties and task descriptions. 

\begin{table*}[ht!]
\centering
\caption{Extracted Dataset Information}
\begin{tabular}{l|cccccc}
\toprule
Dataset & Nodes & Edges & Edge Categories & Timestamps & Domain & Bipartite Graph \\
\midrule
Enron & 42,711 & 797,907 & 10 & 1,006 & E-mail & No \\
GDELT & 6,786 & 1,339,245 & 237 & 2,591 & Knowledge graph & No \\
ICEWS1819 & 31,796 & 1,100,071 & 266 & 730 & Knowledge graph & No \\
Googlemap\_CT & 674,248 & 1,497,006 & 2 & 4,972 & E-commerce & Yes \\
Stack\_elec & 397,702 & 1,262,225 & 2 & 5,224 & Multi-round dialogue & Yes \\
\bottomrule
\end{tabular}
\end{table*}

\section{Implementation Details}
\label{app:details}
Our experiments are conducted on a machine with 2 NVIDIA A100, Intel Xeon CPU (2.30 GHz), and 512GB of RAM.

We rely on the AgentScope framework~\cite{agentscope} to implement GAD and other LLM-based predictors.
We use Llama-Factory~\cite{zheng2024llamafactory} to implement parameter-efficient fine-tuning on Llama.

\section{Complexity Analysis}
In Tables \ref{tab:gnn_cost} and \ref{tab:gad_cost}, we present the time costs for future link prediction using GNNs and GAD on the Enron dataset. 
For GAD, we show the time taken to generate outputs using GPT-4o-mini as the backbone through API calls.

It is evident that for GNN methods, the training time is significantly high, and retraining is required for different tasks and datasets. 
In contrast, for GAD, the overhead before inference is minimal, and the model's knowledge supports multitasking, making it easy to update and reuse within a single dataset.

Regarding inference, GNN methods are accelerated through batching (batch size = 256), resulting in rapid prediction times for individual samples. 
In this work, we focus on having the LLM output results for one sample at a time and do not showcase local acceleration.
With local deployment of large models and parallel acceleration, the GAD framework can achieve higher inference efficiency.

\begin{table}[ht]
\centering
\caption{Time cost for GNN models during training and inference.}
\label{tab:gnn_cost}
\begin{tabular}{@{}lcc@{}}
\toprule
\textbf{GNN Model}        & \textbf{Training Time (s)} & \textbf{Inference Time (per edge) (s)} \\ \midrule
CAWN                      & 157,527                    & -                                     \\
TGAT                      & 87,506                     & -                                     \\
DygFormer                 & 39,398                     & 0.01                                  \\
GraphMixer                & 18,544                     & 0.005                                 \\
TCL                       & 40,405                     & 0.005                                    \\ \bottomrule
\end{tabular}
\end{table}

\begin{table}[ht]
\centering
\caption{Time cost for GAD before and during inference.}
\label{tab:gad_cost}
\begin{tabular}{@{}lc@{}}
\toprule
\textbf{Components of GAD}      & \textbf{Time (s)} \\ \midrule
Local Summary (per node)   & 4.8               \\
Global Summary             & 36.0               \\
Reflection                 & 46.0               \\
\midrule
Prediction (per edge)      & 1.2                \\ \bottomrule
\end{tabular}
\end{table}

\section{Extended Task Description}
\label{app:task_description}
The details of the task descriptions are:
\begin{itemize}[leftmargin=*]
    \item \textbf{Future Link Prediction}: Given the data of nodes \( u \) and \( v \) from time \( 0 \) to \( T \), the goal is to predict whether there will be a link between \( (u, v) \) at time \( T+1 \) or any later timestamps.
    \item \textbf{Node Retrieval}: For a source node \( u \) and a candidate destination nodes set containing one positive sample and several negative samples, node retrieval aims to rank the nodes in the candidate set based on their likelihood of interacting with \( u \) at timestamp \( T+1 \) based on the information collected from time \( 0 \) to \( T \).
    In this paper, we consider the scenario with one positive sample and 100 negative samples.
    \item \textbf{Future Edge Classification}: Similar to link prediction, this task involves observing the data of nodes \( u \) and \( v \) from time \( 0 \) to \( T \), and predicting the type of the edge between \( (u, v) \) at time \( T+1 \) or later. Note that although there exists an edge text between \( u \) and \( v \) at time \( T+1 \), it is not allowed to be used in this task.
\end{itemize} 

\section{Extended Related Works}
\label{app:related works}
\textbf{LLMs for Static Graphs.} Given the effectiveness of LLMs in processing textual information, a series of research has explored incorporating LLMs into solving graph-level tasks. Following~\cite{LLM_explor_Zhikai_23}, these methods can be broadly categorized into two main approaches: LLMs as enhancers and LLMs as predictors.
In the LLMs as enhancers paradigm, textual attributes are utilized to improve the performance of GNN predictors~\cite{TAPE_24, OneForAll_24}.
For instance, TAPE~\cite{TAPE_24} leverages LLMs to generate enriched node features that assist GNNs in downstream tasks.
Conversely, LLM-as-predictors directly use LLMs as standalone predictors for graph-based tasks. 
These approaches capitalize on the rich textual attributes within graphs and have shown remarkable performance~\cite{InstructGLM_23, GraphGPT_24, LLaGa_24, UniGraph_24, GOFA_24}. 
Notably, they unify diverse downstream tasks under a single framework and offer interpretability for their predictions.
However, existing research has primarily focused on static graphs. 
The exploration of LLMs for dynamic graphs remains largely underdeveloped. \\
\textbf{LLM-empowered Agents for Graphs.}
LLM-empowered agents have demonstrated exceptional performance in reasoning and planning tasks, as seen in frameworks such as MetaGPT~\cite{MetaGPT} and HuggingGPT~\cite{HuggingGPT}. 
For a comprehensive review of related work in this domain, we refer the reader to the survey by~\cite{Agent_Survey}.

While LLM-empowered agents have been extensively explored for textual reasoning, their application to graph-based tasks remains less studied. 
\citet{Graph_Agent} introduce an agent-based framework for capturing long-term memory in knowledge graph reasoning. \cite{GraphAgent_Assist} apply multi-agent systems to collaboratively solve predictive and generative tasks in static graphs.
\cite{GraphAgent-Generation} propose a node-wise agent framework to simulate the generation of dynamic text-attributed graphs. 
\cite{GraphAgent-Reasoner} employ multi-agent systems for effective graph reasoning. However, none of these studies explore the potential of LLMs as predictors for dynamic predictive tasks on DyTAGs.

\section{Pareto Principle in Dynamic Graphs}
\label{app:pareto}
\begin{figure*}[htp]
    \centering
    \begin{minipage}{0.3\textwidth}
        \centering
        \includegraphics[width=\linewidth]{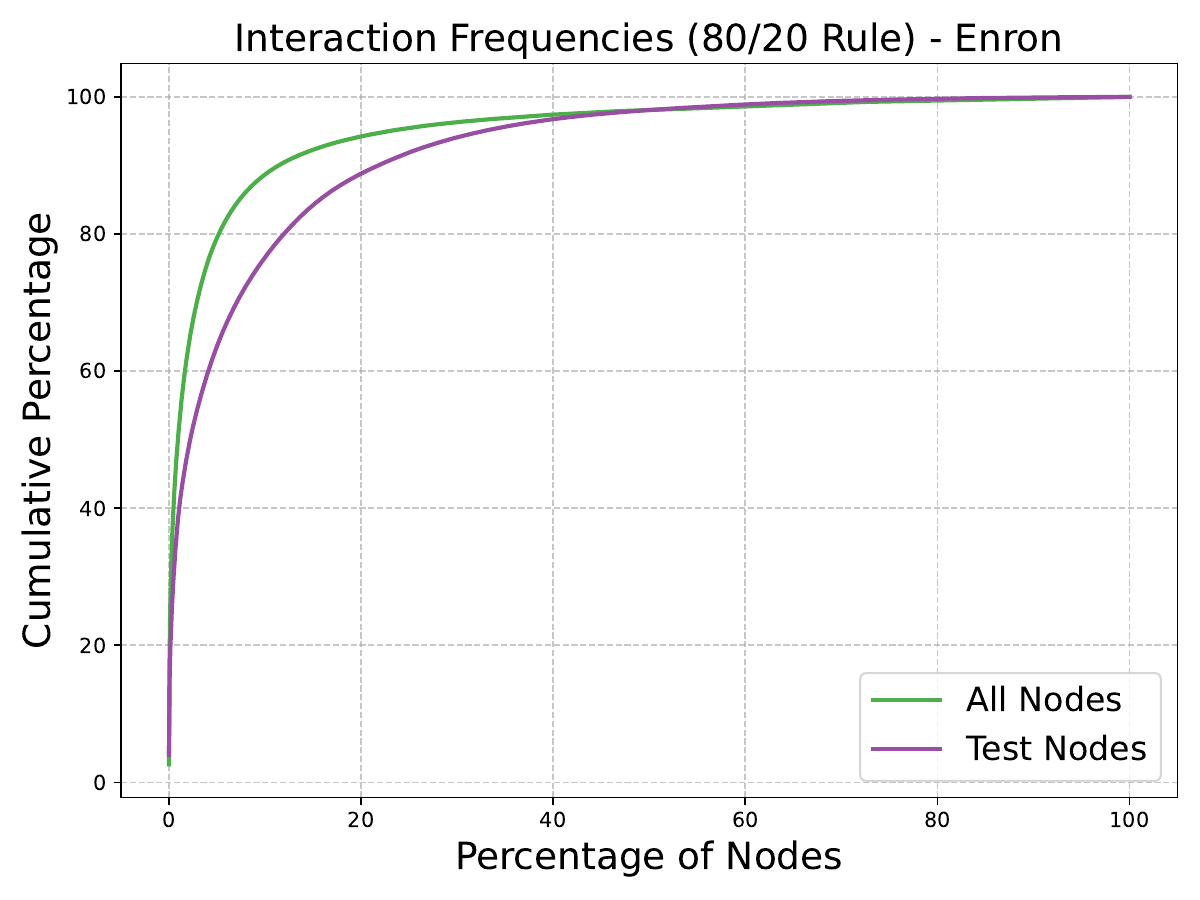}
        \caption*{Enron}
    \end{minipage}\hfill
    \begin{minipage}{0.3\textwidth}
        \centering
        \includegraphics[width=\linewidth]{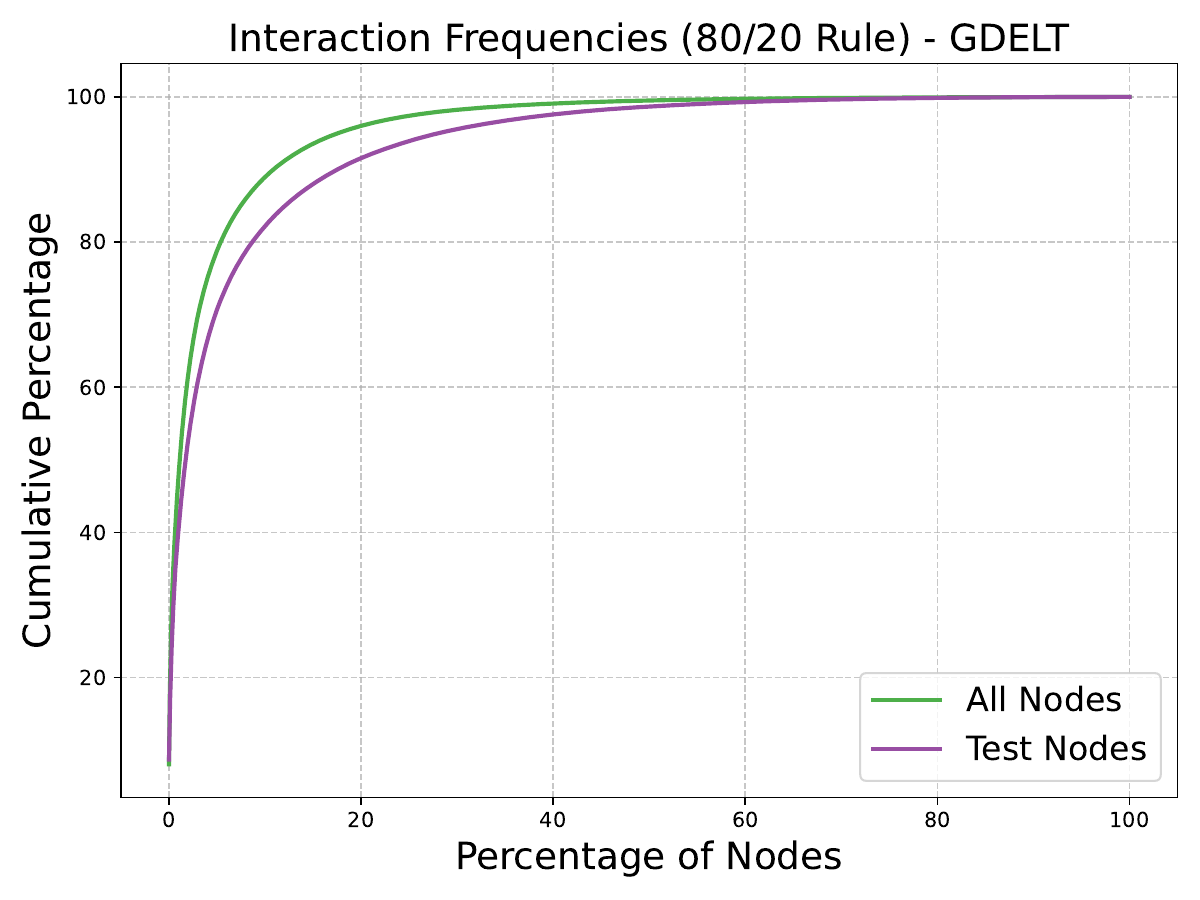}
        \caption*{GDELT}
    \end{minipage}\hfill
    \begin{minipage}{0.3\textwidth}
        \centering
        \includegraphics[width=\linewidth]{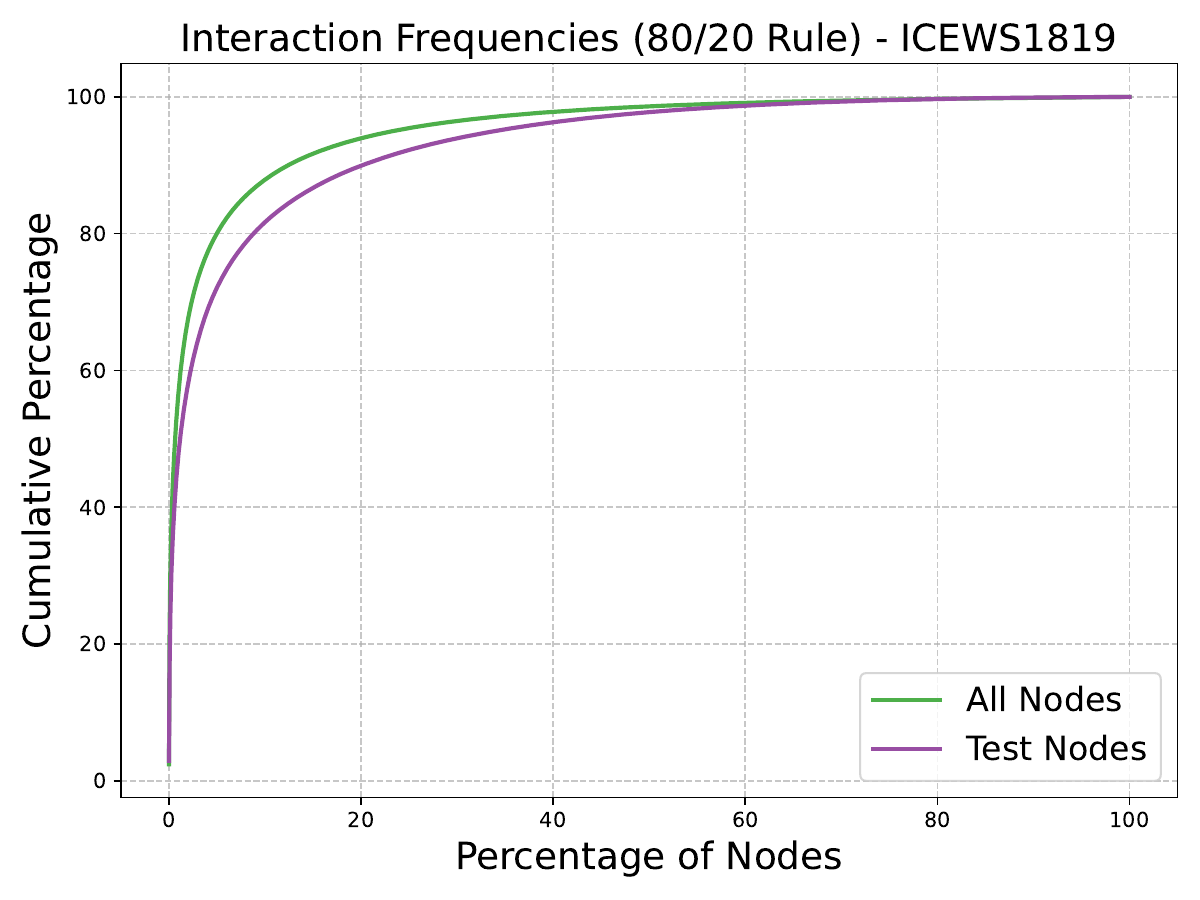}
        \caption*{ICEWS1819}
    \end{minipage}\hfill
    \vskip\baselineskip 
    \begin{minipage}{0.3\textwidth}
        \centering
        \includegraphics[width=\linewidth]{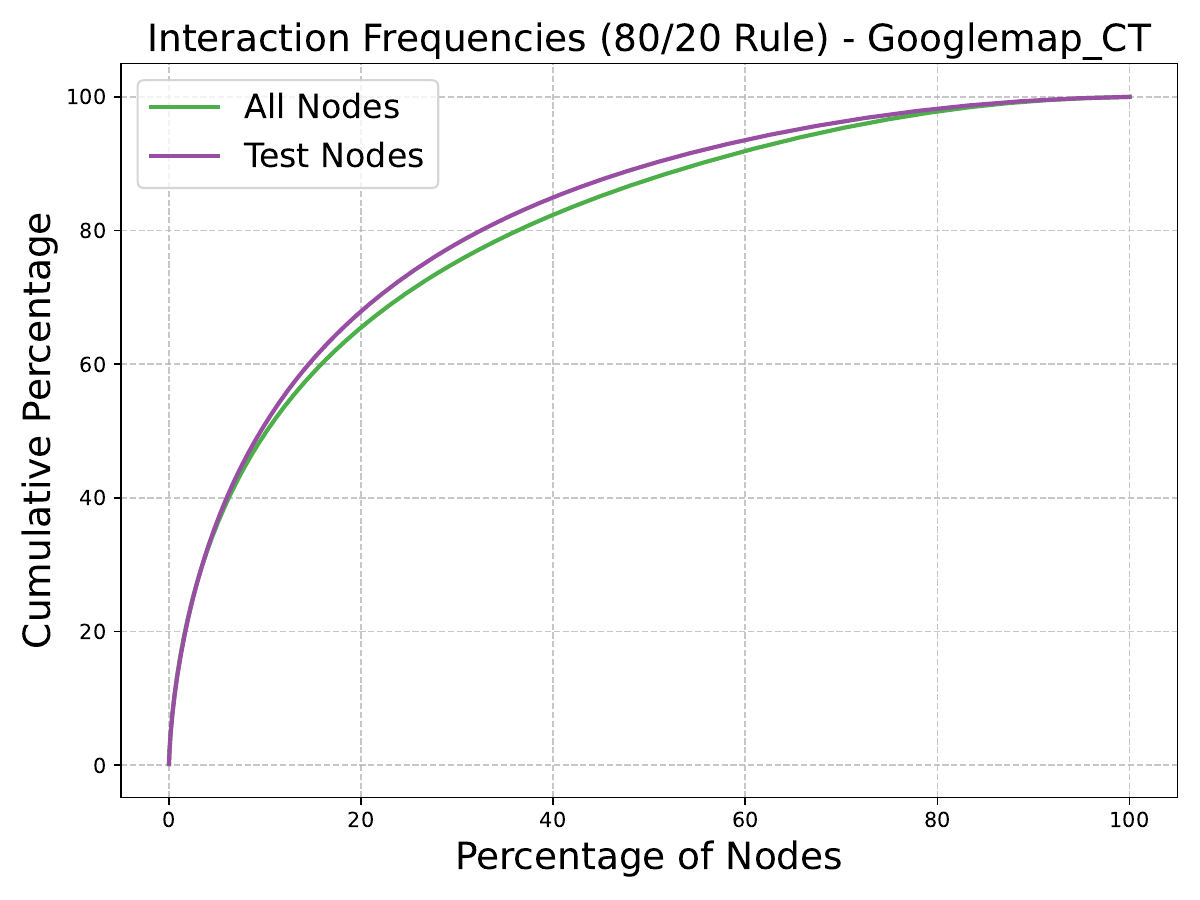}
        \caption*{Googlemap\_CT}
    \end{minipage}\hfill
    \begin{minipage}{0.3\textwidth}
        \centering
        \includegraphics[width=\linewidth]{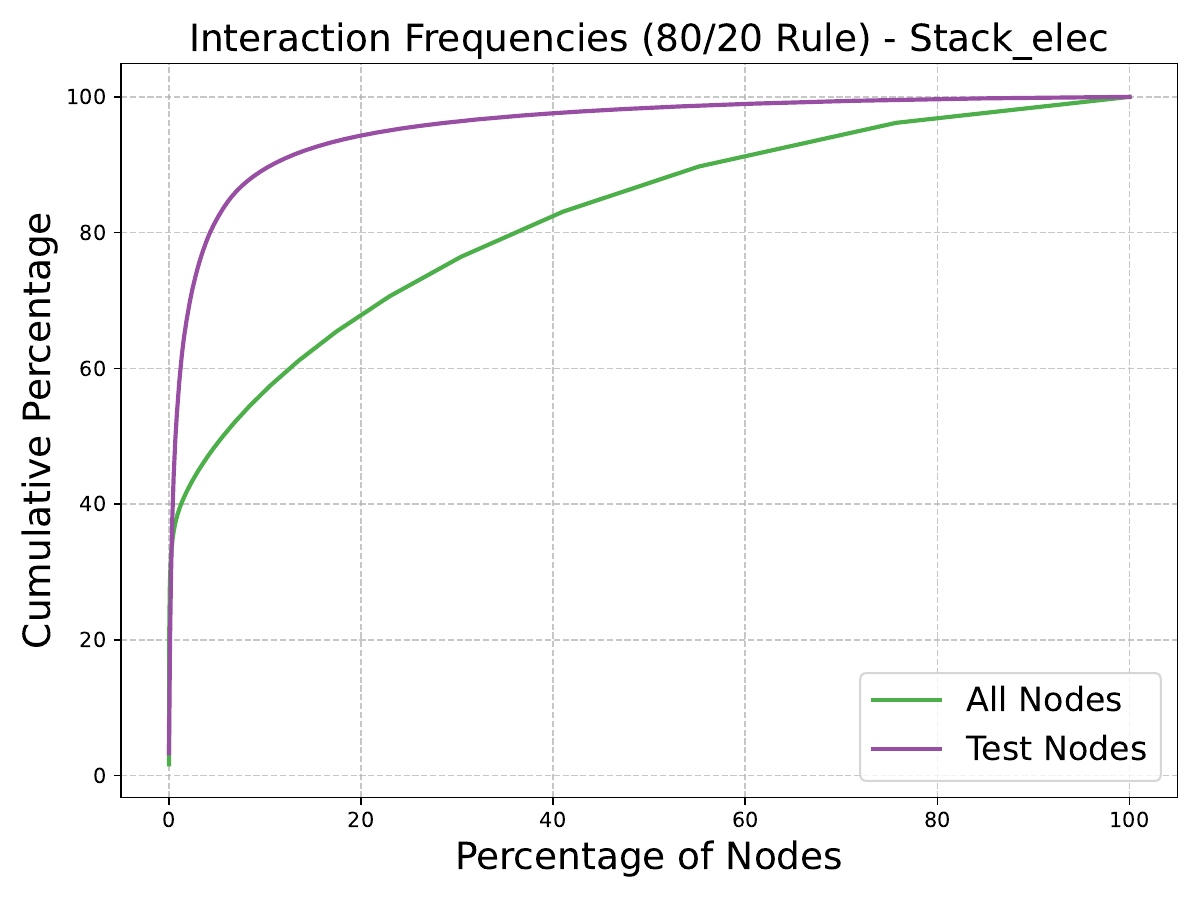}
        \caption*{Stack\_elec}
    \end{minipage}
    \caption{These figures validate the 80/20 law across five datasets. "All nodes" represents the proportion of the most frequent nodes in the training/validation sets that participate in all interactions, while "test nodes" refers to the proportion of these nodes participating in the test phase interactions.}
    \label{fig:dataset_comparison}
\end{figure*}

The Pareto Principle refers to the observation that roughly 80\% of effects come from 20\% of the causes. 
In dynamic graphs, this principle manifests as 20\% of the nodes contributing to 80\% of the interactions. 
In Figure~\ref{fig:dataset_comparison}, we validate this pattern across the five datasets. 
We select the most frequent nodes during the training-validation phase and observe their contribution to interactions in both the entire dataset and the test set. 

As shown in Figure~\ref{fig:dataset_comparison}, the Pareto Principle holds in most datasets. 
Therefore, maintaining local summaries is an efficient and cost-effective approach. 
By focusing on a small subset of important nodes, we can provide significant support for the majority of interactions.

\section{SFT for Node Retrieval}
In the main text, we also include node retrieval samples during the SFT process. However, when we attempted to use the SFT model to improve the predictor's performance, we encountered some obstacles.

To construct the SFT samples, we randomly assigned 5 to 10 negative samples for each positive sample instead of providing all negative samples, given the context length limitation of LLMs. 
The positive samples were assigned a probability of 1.0, while negative samples were assigned a probability of 0.0. However, after SFT, the predictor consistently outputs a non-zero probability for only one sample within the candidate set. If this sample is not the positive one, it results in failures in the hits@3 and hits@10 metrics.

The above results suggest that negative samples must either be appropriately ranked within the fine-tuning samples or that an alternative method is needed to encourage the model to capture uncertainty more effectively. 
Furthermore, the predictor's performance remains constrained by the quality of recall, limiting the potential for SFT to further improve node retrieval.
As a result, further improving performance in node retrieval remains challenging for using LLMs as predictors.

\section{Full Experiment Results}
In Table~\ref{tab:llama_lp} and Table~\ref{tab:llama_ec}, we present full results including Llama-3-8b~\cite{llama3modelcard} as the backbone LLM.
We can see that Deepseek-V3 and GPT4o-mini generally demonstrate better performance than Llama-3-8b, indicating that stronger backbones lead to improved prediction ability in DyTAG tasks.

In Table~\ref{tab:full_results}, we present the full results in the future edge classification task.
We can see that SFT shows a slight improvement compared with the untuned version.

In Table~\ref{tab:node_retrieval}, we present the full results for the node retrieval task, including GAD and GNNs. 
On average, GAD performs comparably to GNNs. However, in the Googlemap\_CT and Stack\_elec datasets, GAD underperforms relative to GNNs due to the absence of helpful metrics. 
Overall, GAD's primary weakness is in Hits@10, indicating that its recall ability requires significant improvement. 
The current rule-based filtering approach sacrifices too much performance for efficiency.

\begin{table*}[ht]
\centering
\caption{Full results of GNNs and single LLM-based predictors in Link Prediction (LP) tasks.
We use accuracy (\%) as the metric.}
\label{tab:llama_lp}
\resizebox{\linewidth}{!}{%
\begin{tabular}{l|rrr|rrrrrrrrr} 
\toprule
\multirow{2}{*}{\textbf{Dataset}} & 
\multicolumn{3}{c|}{\textbf{GNNs}} & 
\multicolumn{3}{c}{\textbf{Text}} &  
\multicolumn{3}{c}{\textbf{Structure}} & 
\multicolumn{3}{c}{\textbf{Structure-FewShot}} \\
\cmidrule(lr){2-4} \cmidrule(lr){5-7}  \cmidrule(lr){8-10} \cmidrule(lr){11-13}
& TCL & GraphMixer & DygFormer & Llama & GPT & Deepseek & Llama & GPT & Deepseek & Llama & GPT & Deepseek \\
\midrule

Enron & 90.56 & 88.08 & \cred{93.18} & 71.84 & 72.59 & 67.28 & 86.17 & 84.80 & 83.37 & 90.15 & 90.78 & \cblue{93.81} \\
GDELT & 90.75 & 89.43 & \cred{91.10} & 55.48 & 76.88 & 72.68 & 80.72 & 80.61 & \cblue{86.11} & 67.70 & 79.33 & 83.39 \\
ICEWS1819 & \cred{95.92} & 94.30 & 95.23 & 68.14 & 78.37 & 83.88 & 89.51 & 90.13 & 91.35 & 90.68 & 87.23 & \cblue{93.00} \\
Googlemap\_CT & \cred{77.51} & 74.00 & 74.23 & 50.18 & \cblue{53.02} & 55.15 & 50.00 & 50.00 & 50.10 & 51.76 & 55.17 & 52.41 \\
Stack\_elec & 91.24 & 91.19 & \cred{91.48} & 53.75 & 51.57 & 51.46  & 50.00 & 50.47 & 51.34 & 43.32 & \cblue{67.66} & 57.03 \\
\bottomrule
\end{tabular}%
}
\end{table*}

\begin{table*}[ht]
\centering
\caption{Full results for Llama-3-8b based predictors on future edge classification task using weighted precision, recall, and F1-score (\%). 
FAIL occurs due to the out of context length issue.}
\label{tab:llama_ec}
\resizebox{\linewidth}{!}{
\begin{tabular}{lc|ccc|cccc}
\toprule
{\textbf{Dataset}} & {\textbf{Metric}} & 
{\textbf{Text}} &  
{\textbf{Structure}} & 
{\textbf{Structure-FewShot}} &
{\textbf{SFT}} &
{\textbf{SFT-all}} &
{\textbf{SFT-Enron}} &
{\textbf{SFT-Google.}} 
\\
\midrule
\multirow{3}{*}{\textbf{Enron}} 
& Precision & 38.32 & 62.02 & 67.45 & 59.38 & 58.94 & - & 61.77 \\
& Recall & 8.34 & 53.35 & 45.02 & 54.52 & 54.59 & - & 56.03 \\
& F1 & 2.72 & 51.92 & 37.46 & 54.30 & 54.16 & - & 55.01 \\
\midrule
\multirow{3}{*}{\textbf{GDELT}} 
& Precision & 2.65 & 7.44 & FAIL & 9.52 & 9.74 & 28.60 & 5.04 \\
& Recall & 2.45 & 10.52 & FAIL & 7.03 & 10.50 & 11.11 & 11.09 \\
& F1 & 1.46 & 5.18 & FAIL & 4.72 & 7.75 & 5.73 & 5.06 \\
\midrule
\multirow{3}{*}{\textbf{ICEWS1819}} 
& Precision & 3.63 & 24.52 & FAIL & 21.32 & 23.66 & 32.74 & 26.84 \\
& Recall & 3.21 & 22.39 & FAIL & 15.68 & 24.12 & 24.18 & 23.82 \\
& F1 & 2.94 & 13.87 & FAIL & 14.22 & 20.89 & 15.82 & 14.40 \\ 
\midrule
\multirow{3}{*}{\textbf{Googlemap\_CT}} 
& Precision & 3.98 & 57.47 & 57.43 & 58.94 & 59.04 & 55.42 & - \\
& Recall & 16.58 & 66.55 & 56.99 & 66.55 & 66.26 & 65.22 & - \\
& F1 & 6.11 & 59.46 & 53.99 & 57.34 & 60.46 & 54.75 & - \\
\midrule
\multirow{3}{*}{\textbf{Stack\_elec}} 
& Precision & 64.23 & 75.65 & 71.05 & 67.46 & 68.58 & 64.37 & 66.47 \\
& Recall & 52.65 & 44.78 & 39.54 & 72.44 &73.61 & 71.78 & 70.67 \\
& F1 & 55.52 & 43.79 & 37.07 & 68.07 & 68.04 & 65.41 & 67.65\\
\bottomrule
\end{tabular}
}
\end{table*}

\begin{table*}[ht]
\centering
\caption{\textbf{Full performance comparison (\%) on future edge classification task} using weighted precision, recall, and F1-score. 
Values are multiplied by 100 for percentage representation. 
{\cred{Red}} indicates the best performance among GNN methods; 
{\cblue{Blue}} denotes the best result in LLM-based predictors.}
\label{tab:full_results}
\resizebox{\linewidth}{!}{
\begin{tabular}{lc|ccc|cccccccc|cc}
\toprule
\multirow{2}{*}{\textbf{Dataset}} & \multirow{2}{*}{\textbf{Metric}} & 
\multicolumn{3}{c|}{\textbf{GNNs}} & 
\multicolumn{2}{c}{\textbf{Text}} & 
\multicolumn{2}{c}{\textbf{Text-FewShot}} & 
\multicolumn{2}{c}{\textbf{Structure}} & 
\multicolumn{2}{c|}{\textbf{Structure-FewShot}} &
\multicolumn{2}{c}{\textbf{GAD}} 
\\
\cmidrule(lr){3-5} \cmidrule(lr){6-7} \cmidrule(lr){8-9} \cmidrule(lr){10-11} \cmidrule(lr){12-13} \cmidrule(lr){14-15}
 & & TCL & GraphMixer & DygFormer & GPT & Deepseek & GPT & Deepseek & GPT & Deepseek & GPT & Deepseek & GPT & Deepseek\\
\midrule
\multirow{3}{*}{\textbf{Enron}} 
& Precision & 12.45 & \cred{50.56} & 47.32 & 38.09 & 40.47 & 43.90 & 41.74 & 55.82 & 59.06 & 55.77 & 58.98 & \cblue{60.24} & 59.85\\
& Recall & 35.28 & \cred{49.49} & 47.55 & 41.19 & 24.95 & 39.04 & 46.48 & 54.12 & 55.28 & 52.97 & \cblue{55.29} & 55.32 & 55.39 \\
& F1 & 18.40 & \cred{48.36} & 46.75 & 31.22 & 28.02 & 31.12 & 43.18 & 53.03 & 53.93 & 52.65 & \cblue{53.97} & 53.74 & 53.87\\
\midrule
\multirow{3}{*}{\textbf{GDELT}} 
& Precision & 00.68 & 10.46 & \cred{16.13} & 09.25 & 06.70 & 08.36 & 06.32 & \cblue{17.45} & 14.96 & 16.91 & 11.58 & 18.51 & 15.19\\
& Recall & 8.24 & 12.94 & 14.74 & 3.85 & 6.70 & 6.41 & 8.43 & 12.08 & 14.57 & 12.50 & 15.00 & 13.89 & 15.17\\
& F1 & 1.26 & 8.92 & \cred{12.15} & 3.44 & 4.44 & 3.86 & 5.54 & 9.44 & \cblue{12.89} & 8.25 & 11.62 & 10.42 & 13.16\\
\midrule
\multirow{3}{*}{\textbf{ICEWS1819}} 
& Precision & 30.37 & 29.68 & \cred{33.21} & 12.16 & 13.30 & 13.98 & 15.21 & 31.20 & 30.22 & \cblue{31.22} & 30.53 & 30.94 & 28.91\\
& Recall & 35.48 & 35.94 & \cred{36.87} & 05.55 & 05.64 & 08.69 & 13.72 & 28.29 & \cblue{33.35} & 29.60 & 33.32 & 31.25 & 32.35\\
& F1 & 29.66 & 29.24 & \cred{31.30} & 6.09 & 4.85 & 7.83 & 10.92 & 24.82 & 29.77 & 26.65 & \cblue{30.21} & 26.41 & 28.59\\
\midrule
\multirow{3}{*}{\textbf{Googlemap\_CT}} 
& Precision & 42.12 & \cred{55.49} & 54.88 & 46.93 & 48.76 & 47.98 & 49.17 & 57.75 & \cblue{58.19} & 55.36 & 57.49 & 55.57 & 56.66\\
& Recall & 64.90 & \cred{65.47} & 65.35 & 13.52 & 8.88 & 19.81 & 36.35 & 63.75 & \cblue{66.16} & 54.01 & 66.35 & 65.16 & 65.63\\
& F1 & 51.09 & 55.17 & \cred{55.18} & 7.69 & 4.16 & 12.53 & 37.11 & 59.42 & \cblue{59.78} & 53.11 & 57.64 & 55.96 & 55.96\\
\midrule
\multirow{3}{*}{\textbf{Stack\_elec}} 
& Precision & 54.32 & \cred{80.62} & 68.43 & 63.74 & 63.50 & 64.32 & 64.06 & 63.97 & 63.59 & \cblue{68.73} & 63.31 & 64.74 & 60.27\\
& Recall & 73.70 & 73.71 & \cred{73.79} & 50.58 & 67.65 & 66.77 & 68.04 & 71.71 & 71.62 & 67.45 & \cblue{72.60} & 71.83 & 69.61\\
& F1 & 62.54 & 62.57 & \cred{65.24} & 53.47 & 65.04 & 65.35 & 65.51 & 65.13 & 64.90 & \cblue{68.03} & 64.16 & 65.48 & 62.98\\
\bottomrule
\end{tabular}
}
\end{table*}

\begin{table*}[ht]
\centering
\caption{Full Node Retrieval performance with performance gaps.
The average gap stands for the average performance gap between GAD and GNNs.
The max gap stands for the performance gap between GAD and the best variants among GNNs.}
\label{tab:node_retrieval}
\begin{tabular}{cc|ccccc|rr}
\toprule
\textbf{Dataset} & \textbf{Metric} & TCL & GraphMixer & DygFormer & GAD\_GPT & GAD\_Deepseek & Avg. gap & Max gap \\
\midrule
\multirow{3}{*}{Enron} & Hits@1 & 46.46 & 42.90 & 76.04 & 72.79 & 72.21 & -17.37 & 3.54 \\ 
 & Hits@3 & 72.26 & 66.19 & 88.27 & 83.72 & 84.28 & -8.43 & 4.27 \\ 
 & Hits@10 & 87.76 & 81.02 & 96.98 & 87.25 & 87.64 & 1.14 & 9.54 \\ 
\midrule
\multirow{3}{*}{GDELT} & Hits@1 & 44.53 & 39.40 & 46.25 & 50.01 & 50.03 & -6.63 & -3.77 \\ 
 & Hits@3 & 68.08 & 65.14 & 71.59 & 72.22 & 71.90 & -3.79 & -0.47 \\ 
 & Hits@10 & 90.86 & 88.34 & 91.32 & 90.09 & 90.62 & -0.18 & 0.97 \\ 
 \midrule
\multirow{3}{*}{ICEWS1819} & Hits@1 & 81.32 & 74.48 & 80.99 & 76.18 & 76.52 & 2.58 & 4.97 \\ 
 & Hits@3 & 93.60 & 89.44 & 92.49 & 87.68 & 87.52 & 4.24 & 6.00 \\ 
 & Hits@10 & 97.91 & 96.24 & 97.31 & 93.00 & 91.68 & 4.81 & 5.57 \\ 
 \midrule
\multirow{3}{*}{Googlemap\_CT} & Hits@1 & 15.84 & 10.83 & 13.86 & 8.85 & 10.27 & 3.95 & 6.28 \\ 
 & Hits@3 & 31.87 & 22.59 & 28.63 & 18.54 & 20.69 & 8.08 & 12.26 \\ 
 & Hits@10 & 57.67 & 43.41 & 51.51 & 40.49 & 41.70 & 9.77 & 16.58 \\ 
 \midrule
\multirow{3}{*}{Stack\_elec} & Hits@1 & 7.63 & 26.00 & 8.22 & 3.58 & 4.28 & 10.02 & 22.07 \\ 
 & Hits@3 & 21.89 & 44.12 & 11.56 & 11.75 & 13.44 & 13.26 & 31.53 \\ 
 & Hits@10 & 56.65 & 80.73 & 15.89 & 41.48 & 42.12 & 9.29 & 38.93 \\ 
 \bottomrule
\end{tabular}
\end{table*}

\section{Examples of Agent Output} \label{app:examples}

\subsection{Global Summary}
Here, we present the complete global knowledge generated for the Enron dataset.  
Thresholds are provided as indicators, and explanations are included to enhance interpretability.  

\bigskip
\noindent
\textbf{Global Link Summary}
\begin{itemize}[leftmargin=*]
    \item \textbf{Textual Analysis}
    \begin{itemize}[leftmargin=2em]
        \item \textbf{Significance:} Not Relevant
        \item \textbf{Reason:} The email addresses in both positive and negative pairs are generic identifiers. They do not convey semantic differences or indicate a stronger relationship in one pair over the other.
        \item \textbf{Explanation:} Since email addresses serve solely as identifiers without inherent semantic meaning, they offer no valuable clues to distinguish between positive and negative relationships.
    \end{itemize}
    
    \item \textbf{Structural Analysis}
    \begin{itemize}[leftmargin=2em]
        \item \textbf{Historical Interaction}
        \begin{itemize}[leftmargin=2em]
            \item \textbf{Significance:} Extremely Significant
            \item \textbf{Explanation:} Positive samples typically exhibit historical interactions, whereas negative samples do not. A historical interaction count greater than 3 strongly indicates a positive sample (65\% of positives), while a count of 0 is characteristic of negatives.
            \item \textbf{Positive Indicator:} Historical interaction count $>$ 3
            \item \textbf{Negative Indicator:} Historical interaction count = 0
        \end{itemize}
        
        \item \textbf{Common Neighbors}
        \begin{itemize}[leftmargin=2em]
            \item \textbf{Significance:} Extremely Significant
            \item \textbf{Explanation:} Positive samples tend to share more common neighbors than negatives. More than 5 common neighbors almost certainly indicates a positive relationship (49\% of positives vs.\ 1\% of negatives), while 0 common neighbors is indicative of a negative sample.
            \item \textbf{Positive Indicator:} Common neighbors $>$ 5
            \item \textbf{Negative Indicator:} Common neighbors = 0
        \end{itemize}
        
        \item \textbf{Destination Node Frequency}
        \begin{itemize}[leftmargin=2em]
            \item \textbf{Significance:} Maybe Related
            \item \textbf{Explanation:} Although a Destination Node Frequency greater than 5 slightly favors positive samples (86\% of positives versus 24\% of negatives), there is considerable overlap between the two classes.
            \item \textbf{Positive Indicator:} Destination Node Frequency $>$ 5
            \item \textbf{Negative Indicator:} Destination Node Frequency = 0
        \end{itemize}
        
        \item \textbf{Overall Structural Indicators}
        \begin{itemize}[leftmargin=2em]
            \item \textbf{Positive Indicator:} Historical interaction count $>$ 3 \textbf{or} Common neighbors $>$ 5
            \item \textbf{Negative Indicator:} Historical interaction count = 0 \textbf{and} Common neighbors = 0
        \end{itemize}
        
        \item \textbf{Structure Rules and Report:} Negative samples exhibit no historical interactions (100\% with count = 0) and almost no common neighbors (94\% with count = 0). In contrast, positive samples show frequent historical interactions (72\% with count $>$ 0) and have a higher number of common neighbors (77\% with count $>$ 0). These combined metrics provide a robust framework for classifying samples with high confidence.
    \end{itemize}
\end{itemize}

\bigskip
\noindent
\textbf{Global Edge Label Summary}
\begin{itemize}[leftmargin=2em]
    \item \textbf{Node Text Analysis}
    \begin{itemize}[leftmargin=2em]
        \item \textbf{Significance:} Maybe Related
        \item \textbf{Reason:} While the email domains (e.g., \texttt{enron.com}, \texttt{caiso.com}) offer some organizational context, they do not provide sufficient semantic detail to reliably predict edge labels.
        \item \textbf{Explanation:} The domain-specific patterns in email addresses are too generic to distinguish between labels such as \texttt{notes\_inbox}, \texttt{personal}, or \texttt{deal\_communication}.
    \end{itemize}
    
    \item \textbf{Edge Text Analysis}
    \begin{itemize}[leftmargin=2em]
        \item \textbf{Significance:} Extremely Significant
        \item \textbf{Reason:} The content of the emails contains strong semantic cues that correlate directly with the assigned edge labels.
        \item \textbf{Explanation:} For instance, emails labeled \texttt{notes\_inbox} often include formal language and document references, \texttt{personal} emails use casual language with non-work topics, and \texttt{deal\_communication} emails discuss business negotiations.
    \end{itemize}
    
    \item \textbf{Edge Label Summary}
    \begin{itemize}[leftmargin=2em]
        \item \textbf{Significance:} Extremely Significant
        \item \textbf{Reason:} Historical reoccurrence patterns for edge labels (across source nodes, destination nodes, and node pairs) show a dominant frequency—over 50\% of communications follow the most frequent label.
        \item \textbf{Explanation:} This consistency in communication patterns strongly suggests that future edge labels will mirror historical trends, making the most frequent historical label a reliable predictor.
    \end{itemize}
\end{itemize}

\subsection{Thresholds}
The generated thresholds for node retrieval across the five datasets are as follows:  
\begin{itemize}[leftmargin=*]
    \item \textbf{Enron:} \( \text{HI} < 1, \text{CN} < 1 \)  
    \item \textbf{GDELT:} \( \text{HI} < 1, \text{CN} < 2 \)  
    \item \textbf{ICEWS1819:} \( \text{HI} < 1, \text{CN} < 1 \)  
    \item \textbf{GoogleMap\_CT:} \( \text{DNF} < 1 \)  
    \item \textbf{Stack\_Elec:} \( \text{HI} < 1 \) and \( \text{DNF} > 5 \)  
\end{itemize}  

\subsection{Reflection Knowledge}
Among the five datasets, only the reflection knowledge in GDELT, ICEWS1819, and GoogleMap\_CT is marked as significant.  
Below, we present the generated reflection knowledge for GDELT.  

GDLET: ``When historical interaction count is 0 and common neighbors are high, then prioritize textual analysis to avoid false positives due to lack of contextual relevance.''

\subsection{Local Knowledge}
In the node-wise knowledge example, we present the generated content of Node 1 in the Enron dataset.  
Note that, we include edge text preference, although it is not used in our tasks.  
Explanations are also provided to enhance interpretability.  

\begin{itemize}[leftmargin=*]
    \item \textbf{Node Description}: The node represents an 'employee', likely involved in regulatory, legal, or energy-related communications within Enron, as evidenced by frequent discussions about California Senate Bill No. 1 and energy policy.
    \item \textbf{Neighbor Preference}: The node prefers neighbors who are colleagues within Enron or external contacts involved in regulatory or energy-related matters, such as \\schwabalerts.marketupdates@schwab.com.
    \item \textbf{Edge Text Preference}: The node prefers edge texts that include detailed discussions on regulatory updates, energy policies, and scheduling, often involving attachments or formal communication.
    \item \textbf{Edge Label Preference}: The node prefers edges with labels 'notes\_inbox' (31.26\%), 'deal\_communication' (30.26\%), and '\_americas' (29.98\%), indicating a focus on internal updates, deal-related communication, and regional (Americas) activities.
    \item \textbf{Explanation}: The node's behavior aligns with an employee deeply involved in regulatory and energy-related matters, particularly within the context of California's energy policies. The high frequency of 'notes\_inbox' and 'deal\_communication' labels suggests a role that requires staying updated on internal communications and deal-related activities. The preference for edge texts involving regulatory updates and formal communication further supports this role. The inclusion of external contacts like schwabalerts.marketupdates@schwab.com indicates engagement with market updates, reinforcing the node's focus on energy and regulatory affairs. The structural preference for 'notes\_inbox' and 'deal\_communication' highlights the importance of internal and deal-specific communication in this role.
    \item \textbf{Structural Preference}: The node prefers well-established and highly connected neighbors, indicating a tendency to engage with influential and active entities within the network.
\end{itemize}

\section{Full Prompts} \label{app:prompts}
In Table~\ref{prompts:control}, we list the prompt for the Initial Agent.
In Table~\ref{prompts:general}, we list the general prompt blocks that are shared in the following Agents.

In Table~\ref{prompts:predictor}, we list the prompts that are used by single LLMs as predictors in Section~\ref{Sec:given text}.

In Table~\ref{prompts:summary_0}, Table~\ref{prompts:summary_1},
Table~\ref{prompts:summary_2}, and
Table~\ref{prompts:summary_3}, we list the prompts for Global Summary Agents.
In Table~\ref{prompts:local_1}, we list the prompts for Local Summary Agents.
In Table~\ref{prompts:reflection}, we list the prompts for the Knowledge Reflection Agent.
In Table~\ref{prompts:GAD}, we list the prompt for the predictor of GAD.
The predictor largely follows the design of single LLMs as predictors, with the including of summary and reflections.

\begin{table*}[ht]
  \caption{The prompt template for the Initial Agent}
  \label{prompts:control}
  \begin{tcolorbox}[colback=blue!5, 
  colframe=black,
  ]
SYSTEM: You are an expert agent specialized in processing dynamic graph datasets. You will receive descriptions of dynamic graphs and the task. 
  You need to extract the necessary information from the description to solve the task.

INPUT: I will now provide the details of the dynamic graph dataset. Please solve the {\{task\_name\}} task.

Dataset Description.

Content\_hints: {{
      "thought": "What you thought",
      "speak": "what you speak",
      "task\_type": "Specifies the downstream task for the graph analysis (e.g., 'link prediction', 'edge classification').",
      "graph\_type": "Describes the type of graph in use, focusing on its specific domain or context (e.g., 'knowledge graph', 'social network', 'citation network', 'email network').",
      "node\_type": "Defines the types of nodes in the graph (e.g., 'user', 'user' and 'item').",
      "node\_text\_type": "Describes the semantic meaning of the textual content associated with each node type.",
      "edge\_type": "Defines what the relationship between nodes represents.",
      "edge\_text\_type": "Provides the semantic meaning of the textual content representing the relationship between nodes."
  }}

  \end{tcolorbox}
  \end{table*}

\begin{table*}[ht]
  \caption{The prompt template blocks}
  \label{prompts:general}
  \begin{tcolorbox}[colback=blue!5, 
  colframe=black,
  ]
\textbf{Global Description Block}

You are an expert node analyst specializing in analysis within a {\{graph\_type\}} network.
Nodes are {\{node\_type\}} with textual features, where node text means {\{node\_text\_type\}}.
Edges represent {\{edge\_type\}} relationships, with textual features meaning {\{edge\_text\_type\}}.

\textbf{Task Description Block}

Future Link Prediction:

SYSTEM:
For each pair of source and destination nodes, determine whether a link will form between them based on the provided metrics.

INPUT:
Predict the existence of an edge between {\{src\_id\}} and {\{dst\_id\}}. 
Respond with '1' (int) for yes or '0' (int) for no.

Node Retrieval:

SYSTEM:
Assign a probability between 0 and 1 to each provided destination node based on their likelihood of future interaction.

INPUT:
Given the above information, assign a probability between 0 and 1 to each Destination Node ID based on their likelihood of future interaction with the source node. A higher probability indicates a higher likelihood of interaction.
Respond with a JSON object where each key is the Destination Node ID (str), and the value is its probability (float between 0 and 1).

Future Edge Classification:

SYSTEM:
Important: You are only allowed to use the provided classes to classify the edges. Do not introduce any additional classes. 
Use the provided information to make a well-informed classification of the edge. Consider both historical patterns and feature relevance when making your decision.

INPUT:
Predict the class label for the edge between {\{src\_id\}} and {\{dst\_id\}}.
Respond with a JSON object where the key is 'Prediction' and the value is `'edge\_class'` (a string from the provided classes).
Use the original text of class from provided classes. Do not introduce any additional classes or modifications.

\textbf{Example Block}

Except for the Initial Agent, we include an example output format to guide the LLM to output in the right format.

  \end{tcolorbox}
  \end{table*}

\begin{table*}[ht]
  \caption{The prompt template for Predictor Agent}
  \label{prompts:predictor}
  \begin{tcolorbox}[colback=blue!5, 
  colframe=black,
  ]

\textbf{Link Prediction Base Prompt}

SYSTEM: 

{\{Global Description\}}
Your task is to predict whether two nodes will interact based on the following information {\{and several examples (if few\_shot)\}}:

1.  Node Text: 
- Textual information of each node.

2.  Historical Interaction Count: 
- The total number of past interactions the two nodes have had.
- High interaction counts indicate active historical interactions between the two nodes and a higher likelihood of future interactions.

3.  Common Neighbor Count: 
- The number of shared neighbors between the two nodes.
- A higher number of common neighbors suggests a stronger community bond and a higher probability of interaction.

4.  Node-Specific Metrics: 
-  For Source Node: 
    -  Frequency:*Total number of interactions.
    -  Times as Source:*Number of times the node has been the source node in interactions.
    -  Times as Destination:*Number of times the node has been the destination node in interactions.
    -  Average Frequency of Neighbors:*Average number of interactions per historical neighbor, indicating whether the node's neighbors are predominantly new or well-established.

-  For Destination Node: 
    -  Frequency:*Total number of interactions.
    -  Times as Source:*Number of times the node has been the source node in interactions.
    -  Times as Destination:*Number of times the node has been the destination node in interactions.
    -  Average Frequency of Neighbors:*Average number of interactions per historical neighbor, indicating whether the node's neighbors are predominantly new or well-established.

- If the destination node has participated in many interactions overall but has rarely been a destination in past interactions, it is less likely for this pair to form a link. 
- If a node demonstrates a preference for neighbors with high or low interaction frequency, it is likely to exhibit the same preference for forming links with new nodes.

{\{Task Description\}}

INPUT:

 Current Sample: 
 Node Text: 
Source Node text: {\{\}}
Destination Node text: {\{\}}

 Historical Interaction Count: 
The total number of past interactions between Source ID {\{src\_id\}} and Destination ID {\{dst\_id\}}: {\{\}}

 Common Neighbor Count: 
The number of shared neighbors between Source ID {\{src\_id\}} and Destination ID {\{dst\_id\}}: ...

 Node-Specific Metrics: 
 For Source Node ({\{src\_id\}}): 
- Frequency: {\{\}}
- Times as Source: {\{\}} 
- Times as Destination: {\{\}} 
- Average Frequency of Neighbors: {\{\}}

 For Destination Node ({\{dst\_id\}}): 
- Frequency: {\{\}}
- Times as Source: {\{\}} 
- Times as Destination: {\{\}} 
- Average Frequency of Neighbors: {\{\}}

{\{Task Description\}}

\textbf{Node Retrieval Base Prompt}

The prompt for node retrieval is almost the same as those in the link prediction task, except that multiple destination nodes are included.

\textbf{Edge Classification Base Prompt}

SYSTEM: 

{\{Global Description\}}

1. Node Text:
- Textual information of its connected nodes.

2. Node Preferences:
- This dictionary contains the historical edge class distributions for both the source and destination nodes.
- The keys represent class names, and the values indicate the frequency of each class observed in past edges involving the node.
- Higher frequencies suggest a higher likelihood of that class being the predicted class for the edge.

3. Pair Preference:
- This historical classes of edges between the source and destination nodes.
- The keys are the class names, and the values are the frequencies of those classifications in past observations.
- Higher frequencies of a particular class increase the likelihood that the same class will be predicted for the current edge.     

{\{Task Description\}}

INPUT:

 Edge Classes : {\{edge\_class\_names\}}.
Now you need to classify the below edge's class into one of the above classes based on the following information:

 Node Text: 
- Source Node {\{\}}
- Destination Node {\{\}}

 Node Preferences: 
- Source Node ({\{src\_id\}}): {\{ELD\_src\}}
- Destination Node ({\{dst\_id\}}): {\{ELD\_dst\}}

 Pair Preferences: 
- Between Source Node ({\{src\_id\}}) and Destination Node ({\{dst\_id\}}): {{ELD\_pair}}

{\{Task Description\}}

\textbf{Other Variants}

For few-shot variants, examples are attached in the input.
For text variants, metrics except node text are excluded.

  \end{tcolorbox}
  \end{table*}

\begin{table*}[ht]
  \caption{The prompt template for Structural Text Agent in Global Summary Agents}
  \label{prompts:summary_0}
  \begin{tcolorbox}[colback=blue!5, 
  colframe=black,
  ]
SYSTEM:

{\{Global Description\}}
Your task is to analyze the raw textual content of positive and negative pairs to:

1. Determine the significance of the textual information in distinguishing positive and negative pairs based purely on their text, using the following significance levels: 'Extremely Significant', 'Helpful', 'Maybe Related', 'Not Relevant'.

2. Provide key reasons supporting the significance based strictly on the provided text. The explanation should be concise and high-level, avoiding detailed statistics.

3. Explain why any textual trends or patterns emerge within this type of graph based on the provided raw text, without external context.

4. Be cautious and only declare significance when strong textual evidence supports it.

5. Do not rely on any external knowledge, assumptions, or context; only the provided text matters.

Definitions:
   - 'Extremely Significant': The information alone can very well solve the task.
   - 'Helpful': The information aids in solving the task.
   - 'Maybe Related': The information has some relation to the task but is not reliable on its own.
   - 'Not Relevant': The information is unrelated to the task.

INPUT:
{\{text\_samples\}}

Tasks:

1. Assess whether the raw textual content of positive pairs shows clear semantic patterns or higher relevance compared to negative pairs.
   - For example, a positive pair ('china', 'america') is more likely to show a relevant relationship compared to a negative pair ('china', 'cat') purely based on the text itself.
   - Conversely, a positive pair ('user1', 'item1') and a negative pair ('user1', 'item2') may have no meaningful difference in the provided text alone.
   
2. Focus strictly on the provided raw text. You are not allowed to consider any background knowledge or context outside the text.

3. Do not make assumptions about the relationship between the nodes outside of the text given.

4. Look for strong semantic relevance in the text itself rather than relying on superficial textual similarities. The explanation should be concise and high-level, avoiding detailed statistics.

5. Be careful about declaring significance—ensure the text clearly supports the conclusion.

{\{Example Output Format\}}

  \end{tcolorbox}
  \end{table*}

\begin{table*}[ht]
  \caption{The prompt template for Structure Agent in Global Summary Agents}
  \label{prompts:summary_1}
  \begin{tcolorbox}[colback=blue!5, 
  colframe=black,
  ]
SYSTEM:

{\{Global Description\}}
You have distributions of structural statistics for positive and negative samples.
Your goal is to:

1. Determine the significance of each structural metric in distinguishing positive and negative pairs using the following significance levels: 'Extremely Significant', 'Helpful', 'Maybe Related', 'Not Relevant'.

2. Provide explanations for each metric that include numerical guidelines for classification.

3. Identify strict global structural rules that distinguish positive from negative samples with detailed explanations.

4. Provide conditions under which a sample can be almost certainly classified as positive or negative based on the structural metrics.
   -  Positive Indicator:*Conditions where a sample is almost certainly positive and impossible to be negative.
   -  Negative Indicator:*Conditions where a sample is almost certainly negative and impossible to be positive.
   
5. Ensure that the distinctions you identify are highly significant and obvious based on the statistical data provided.
Definitions:
   - 'Extremely Significant': The information alone can very well solve the task. It is one of the most important information.
   - 'Helpful': The information largely aids in solving the task.
   - 'Maybe Related': The information has some relation to the task but is not reliable on its own.
   - 'Not Relevant': The information is unrelated to the task.
   
6. Each structural metric must include both positive and negative indicators with numerical thresholds.

7. The overall indicators must combine all metrics to provide a comprehensive rule for classification with numerical guidelines.

INPUT:

Below are the distributions of structural statistics for positive and negative samples:

 Positive sample distribution: 
  - Historical interaction count: {\{distribution\}}
  - Common neighbors: {\{distribution\}}
  - Destination Node Frequency: {\{distribution\}}

 Negative sample distribution: 
  - Historical interaction count: {\{distribution\}}
  - Common neighbors: {\{distribution\}}
  - Destination Node Frequency: {\{distribution\}}

Tasks:
1. Declare the significance of each structural metric in distinguishing positive and negative pairs using the standardized significance levels.
2. Provide strict scenarios of explanations for each metric that include numerical thresholds guiding classification.
   - Example: 'When historical interaction count > 2, it is very likely to be a positive sample; when > 3, it is almost certainly positive.'
   - When will it be a Good example: 97\% negative samples fail to satisfy, but over 70\% positive samples satisfy historical interaction count > 2.
   - When will it be a Bad example: While over 90\% positive samples satisfy historical interaction count > 2, more than 30\% negative samples satisfy as well.
3. Identify strict structural rules that differentiate positive and negative samples with explanations based on the graph type.
4. Describe strict scenarios where the structural metrics make it impossible for a sample to be of the other type, ensuring minimal misclassification.
   - Positive Indicator: Historical interaction count > X (where X is a threshold indicating the impossibility of being negative).
   - Negative Indicator: Historical interaction count < Y (where Y is a threshold indicating impossibility of being positive).
   - Example of Good Positive Indicator: 97\% negative samples fail to satisfy, but over 70\% positive samples satisfy.
   - Example of Bad Positive Indicator: While over 90\% positive samples satisfy, more than 30\% negative samples satisfy as well.
5. Provide overall indicators that combine all metrics to classify samples as positive or negative with high confidence, including numerical guidelines.

{\{Example Output Format\}}

  \end{tcolorbox}
  \end{table*}

\begin{table*}[ht]
  \caption{The prompt template for Edge Text Agent in Global Summary Agents}
  \label{prompts:summary_2}
  \begin{tcolorbox}[colback=blue!5, 
  colframe=black,
  ]
SYSTEM:

{\{Global Description\}}

Your task is to evaluate the effectiveness of using node and edge textual content to predict edge labels.
Specifically, you should:

1. Determine the significance of textual information in predicting edge labels using the following levels: 'Extremely Significant', 'Helpful', 'Maybe Related', 'Not Relevant'.
   - Including both edge text and node text.
   
2. Provide reasons and guidance in detail supporting the significance based on the dataset, and if significant, illustrate how one can use the texts to predict edge labels.

3. Explanation of how to use the text to predict edge labels and explain why these textual trends occur in this type of graph and provide examples if Significant.

4. Ensure strong support is provided before declaring any significance.
Definitions:
   - 'Extremely Significant': The information alone can very well solve the task.
   - 'Helpful': The information aids in solving the task.
   - 'Maybe Related': The information has some relation to the task but is not reliable on its own.
   - 'Not Relevant': The information is unrelated to the task.

INPUT:

{\{Text Samples\}}

Tasks:

1. Analyze both node text and edge text separately:
   a) For node text: Assess whether the textual content of node pairs shows patterns that can predict edge labels.
   b) For edge text: Assess whether the edge text content exhibits patterns that can predict edge labels.
   
2. Focus solely on the semantic relationships or patterns within the provided texts.

3. Do not rely on superficial similarities or external context. Emphasize semantic relevance over superficial textual differences.

4. When providing guidance, solely rely on the text itself, do not rely on external tools.
   - You can provide guidance on the semantical meaning, tone of the text or its alignment with edge labels.
   - You cannot use complex NLP techniques.
   - Concise and effective guidance is welcomed.
   
5. Ensure strong support is provided before declaring any significance.

{\{Example Output Format\}}

  \end{tcolorbox}
  \end{table*}

\begin{table*}[ht]
  \caption{The prompt template for ELD Agent in Global Summary Agents}
  \label{prompts:summary_3}
  \begin{tcolorbox}[colback=blue!5, 
  colframe=black,
  ]
SYSTEM:

{\{Global Description\}}

You have the reoccurrence distributions of edge label preferences based on historical interactions for source nodes, destination nodes, and node pairs.  
   - The distribution means the likelihood of each edge label to occur in the future, ordered from the most frequent to the least frequent.  
   - The first edge label is the most frequent edge label in the historical data.  
   - The last edge label is the least frequent edge label in the historical data.  
The preference indicates the potential of each edge label observed in the historical data to occur in the future.  
   - For example, if a source node has a high preference for the first label, it suggests that edge labels are likely to be similar to the past.  
   - If a source node has a significantly low preference for the first label, it suggests that edge labels are likely to be different from the past.  
Your goal is to provide direct guidance on using these historical edge label preferences to predict future edge labels.  
   - For example, if the most frequent edge label is likely to occur, you can include 'Use the most frequent historical edge label for predicting future edge labels' in the guidance.  

Specifically, you should:  
1. Explain the significance of the provided edge label distributions using the following levels: 'Extremely Significant', 'Helpful', 'Maybe Related', 'Not Relevant'.  
2. Provide guidance on how historical edge labels can be used to predict future edge labels.  
3. If the correlation is extremely strong and reliable, provide guidance on using historical labels for prediction.  
4. If the correlation is weak or uncertain, indicate that it's Not Significant for reliable prediction.

INPUT:

Below are the distributions of edge label reoccurrence based on historical interactions for positive samples:

Source node historical edge label reoccurrence distribution: {\{distribution\}}
Destination node historical edge label reoccurrence distribution: {\{distribution\}}
Pair node historical edge label reoccurrence distribution: {\{distribution\}}

Analyze whether historical source/destination/pair-wise edge label reoccurrence patterns can predict future edge labels.  
Provide a high-level and direct guidance in predicting future edge labels if patterns exist.

{\{Example Output Format\}}
  \end{tcolorbox}
  \end{table*}

\begin{table*}[ht]
  \caption{The prompt template for Local Summary Agent}
  \label{prompts:local_1}
  \begin{tcolorbox}[colback=blue!5, 
  colframe=black,
  ]
\textbf{Text Preference}

SYSTEM:

{\{Global Description\}}

Your task is to generate a comprehensive profile for a node that accurately reflects its textual characteristics, interaction and neighbor preferences, as well as structural features.
The profile should include the following keys:
    1. Node Description
    2. Neighbor Preference
    3. Edge Text Preference
    4. Edge Label Preference
    5. Explanation

Ensure the output is in JSON format with the exact keys listed above.
For keys 1 to 4, use concise, summary language that expresses preferences.
For the 'Explanation' key, provide a detailed explanation with data support.
If any part is difficult to summarize due to insufficient evidence or data, or if the conclusion is not significant, mark it as 'Not Significant'.
Provide detailed analysis for the 'Explanation' key.

INPUT:

Here is the textual information of the node:  
{\{text\_samples\}}

{\{edge\_label\_explanation\}}

Generate the following details in the exact format as the examples provided below:

1.  Node Description: Summarize the type and characteristics of the node based on its text, using concise language.  

2.  Neighbor Preference: Summarize the node's preferences for types of neighbors based on their texts, using preference-based language.  

3.  Edge Text Preference: Summarize the node's preferences for types of edge texts using preference-based language.  

4.  Edge Label Preference: Summarize the node's preferences for edge labels using the provided edge label distribution.  
   - Its label preference should be inferred solely based on the distribution, not text.  
   
5.  Explanation: Provide a detailed explanation that combines real-world node behavior characteristics with the above details, using data support.  

If any part is difficult to summarize due to insufficient evidence or data, or if the conclusion is not significant, mark it as 'Not Significant'.  
Only if the structural information has significant characteristics should you provide structural Preference.
Ensure that the output strictly follows the example output formats provided below.

{\{Example Output Format\}}

\textbf{Structural Preference}

SYSTEM:

{\{Global Description\}}

Your task is to generate a comprehensive structural preference summary for a node based on its structural features.
The summary should include the following key:
1. Structural Preference

Ensure the output is in JSON format with the exact key listed above.
For the 'Structural Preference' key, use concise, summary language to describe the node's overall structural preferences.
Do not include numerical results. Instead, abstractly summarize characteristics such as preferring well-established, highly connected, active, or new nodes.
If the node does not have a clear structural preference, mark it as "Not Significant".

  \end{tcolorbox}
  \end{table*}

\begin{table*}[ht]
  \caption{The prompt template for Knowledge Reflection Agent}
  \label{prompts:reflection}
  \begin{tcolorbox}[colback=blue!5, 
  colframe=black,
  ]
\textbf{Text Preference}

SYSTEM:

You are a reflection agent responsible for identifying significant missing structural knowledge in the global summary used for link prediction. Your task is to analyze false positive samples and their rate to determine if additional structural insights are needed.

-  INPUT Analysis: 
    - Accuracy: Primary indicator of current performance
    - False Positive Samples: Detailed examples of incorrect positive predictions
    - Global Knowledge: Current understanding of the link prediction task

-  Decision Making: 
    - If Accuracy is high
        - Automatically classify as "Not Significant"
        - No complementation needed
    - If Accuracy is low:
        - Analyze patterns in false positive samples
        - Look for consistent structural patterns that led to incorrect predictions
        - If clear pattern exists, provide complementary knowledge
        - If no clear pattern, mark as "Not Significant"

-  Requirements: 
    - Examine the false positive error samples to identify structural patterns indicating missing knowledge.
    - If false positives indicate incomplete global summary, provide complementary structural insights.
    - If false positives are minimal or do not suggest missing structural knowledge, mark as "Not Significant."
    - Emphasize both structural and textual metrics, with a focus on structural aspects!

{\{Example Output Format\}}

INPUT:

Global Knowledge: {{Global Summary}}

Accuracy: {{Accuracy}}
False Positive Samples: {{Error Samples}}
Review the false positive samples and rate to determine if there is significant missing structural knowledge. 
If significant, provide a single sentence complementation using 'When..., then...' format to address the pattern in false positives.

{\{Example Output Format\}}

  \end{tcolorbox}
  \end{table*}

\begin{table*}[ht]
  \caption{The prompt template for GAD Predictor Agent}
  \label{prompts:GAD}
  \begin{tcolorbox}[colback=blue!5, 
  colframe=black,
  ]
\textbf{Text Preference}

SYSTEM: 

{\{Global Description\}}

{\{Global Summary\}} + {\{Knowledge Reflection\}}

{\{Task Description\}}

INPUT:

The INPUT to GAD is similar to a single LLM with Structural Prompt, except that:
A local Summary is provided if available, and only significant metrics are included.

\end{tcolorbox}
\end{table*}

\end{document}